\documentclass{article}
\usepackage{ijcai19}

\usepackage{times}
\usepackage{soul}
\usepackage{url}
\usepackage[hidelinks]{hyperref}
\usepackage[utf8]{inputenc}
\usepackage[small]{caption}
\usepackage{graphicx}
\usepackage{amsmath}
\usepackage{booktabs}
\usepackage{algorithm}

\usepackage{subfig}
\usepackage{amssymb}
\usepackage{subfloat}
\usepackage{algorithmicx}
\usepackage{algpseudocode}
\usepackage[OT1]{fontenc}
\usepackage{natbib}
\hypersetup{
  colorlinks,
  citecolor=blue,
  linkcolor=black,
  urlcolor=black}

\newenvironment{sequation}{\small\begin{equation}}{\end{equation}}

\newcommand{\mainsectionstyle}{%
  \renewcommand{\@secnumfont}{\bfseries}
  \renewcommand\subsubsection{\@startsection{section}{2}%
    \z@{.5\linespacing\@plus.7\linespacing}{-.5em}%
    {\normalfont\bfseries}}%
}

\title{Understanding Memory Modules on Learning Simple Algorithms}

\author{
Kexin Wang$^{1, 2}$
\and
Yu Zhou$^{1, 2}$\and
Shaonan Wang$^{1, 2}$\and
Jiajun Zhang$^{1, 2}$\And
Chengqing Zong$^{1, 2, 3}$\\
\affiliations
$^1$National Laboratory of Pattern Recognition, CASIA, Beijing, China\\
$^2$University of Chinese Academy of Sciences, Beijing, China\\
$^3$CAS Center for Excellence in Brain Science and Intelligence Technology, Beijing, China\\
\emails
\{kexin.wang, yzhou, shaonan.wang, jjzhang, cqzong\}@nlpr.ia.ac.cn}
\begin{document}

\maketitle

\begin{abstract}
 Recent work has shown that memory modules are crucial for the generalization ability of neural networks on learning simple algorithms. However, we still have little understanding of the working mechanism of memory modules. To alleviate this problem, we apply a two-step analysis pipeline consisting of first inferring hypothesis about what strategy the model has learned according to visualization and then verify it by a novel proposed qualitative analysis method based on dimension reduction. Using this method, we have analyzed two popular memory-augmented neural networks, neural Turing machine and stack-augmented neural network on two simple algorithm tasks including reversing a random sequence and evaluation of arithmetic expressions. Results have shown that on the former task both models can learn to generalize and on the latter task only the stack-augmented model can do so. We show that different strategies are learned by the models, in which  specific categories of input are monitored and different policies are made based on that to change the memory.

 % To the best of our knowledge, this is the first time that hypothesis-verification pipelines about what strategy is induced are carried out for memory augmented neural networks.
\end{abstract}

\section{Introduction}
The generalization ability of neural networks is the most important criterion to determine whether they are powerful or not. Recent work on memory-augmented neural networks (MANN) has shown promising results about generalizing perfectly on simple-algorithm tasks~\citep{grefenstette2015learning, joulin2015inferring,graves2016hybrid, rae2016scaling, gulcehre2017memory}. However, understanding the working mechanism of memory modules has not been well studied. Existing work has stopped at conjecturing the underlying learned strategies based on shallow, single-case visualization without any further verification. This raises concern of the lack of interpretability and hinders us from designing better memory modules. Currently, there are two main difficulties for interpreting the black-box memory-modules. 

Firstly, the diversity among the different MANNs makes it hard to isolate the functions of memory-modules. Consequently, we cannot focus on the decisive parts of the models, as the different designs make the comparison work hard to carry out. 
% Secondly, it is difficult to construct a proper dataset to probe the strategy learned by models on latent-tree structures. There are a few existing related studies. However, they propose data with simple latent structures, such as sequences $a^nb^nc^n$~\citep{joulin2015inferring}, and repeated sequences~\citep{gulcehre2017memory}.
Secondly, to interpret MANNs is challenging. Although interpretation methods for RNN models are well studied, little attention has been paid on that for MANNs.

To solve the above problems, we formalize a unified framework for MANNs with different implementations of memory modules, which makes comparison among different memory modules feasible. % Then, we design an evaluation task of \emph{modulo-$10$ arithmetic expressions}, \emph{M10AE}, in which modulo-$10$ refers to a modulo-$10$ operator. \emph{M10AE} simulates the tree structures of natural languages and also the small vocabulary keeps the analysis work tractable. 
Then, we propose a novel qualitative analysis method based on dimension reduction for interpreting memory cells by verifying hypotheses. We implement neural Turing machine and stack-augmented neural network under the unified framework and carry out detailed analysis on two algorithm tasks consisting of reversing a random sequence and evaluating arithmetic expressions to show the effectiveness of our proposed analysis method.

The experiment and analysis have shown that the external memory can compensate for the need for storing intermediate results to travel along the dependency path. Specifically, neural Turing machine generalizes well on mirror task, and stack-augmented neural network generalizes well on reversing a random sequence and evaluation of arithmetic expressions.
To summarize, our main contributions are as follows:
\begin{itemize}
  \item We generalize different MANNs with a unified framework, which ensures the fairness of comparing different memory mechanisms.
  \item We propose a novel analysis method for memory cells. By applying our analysis method, we infer and verify the hypotheses about what strategy has been learned by the model that can generalize well.
\end{itemize}

\section{Our unified MANN framework}
\begin{figure}[!htbp]
  \centering
  \includegraphics[width=.35\textwidth]{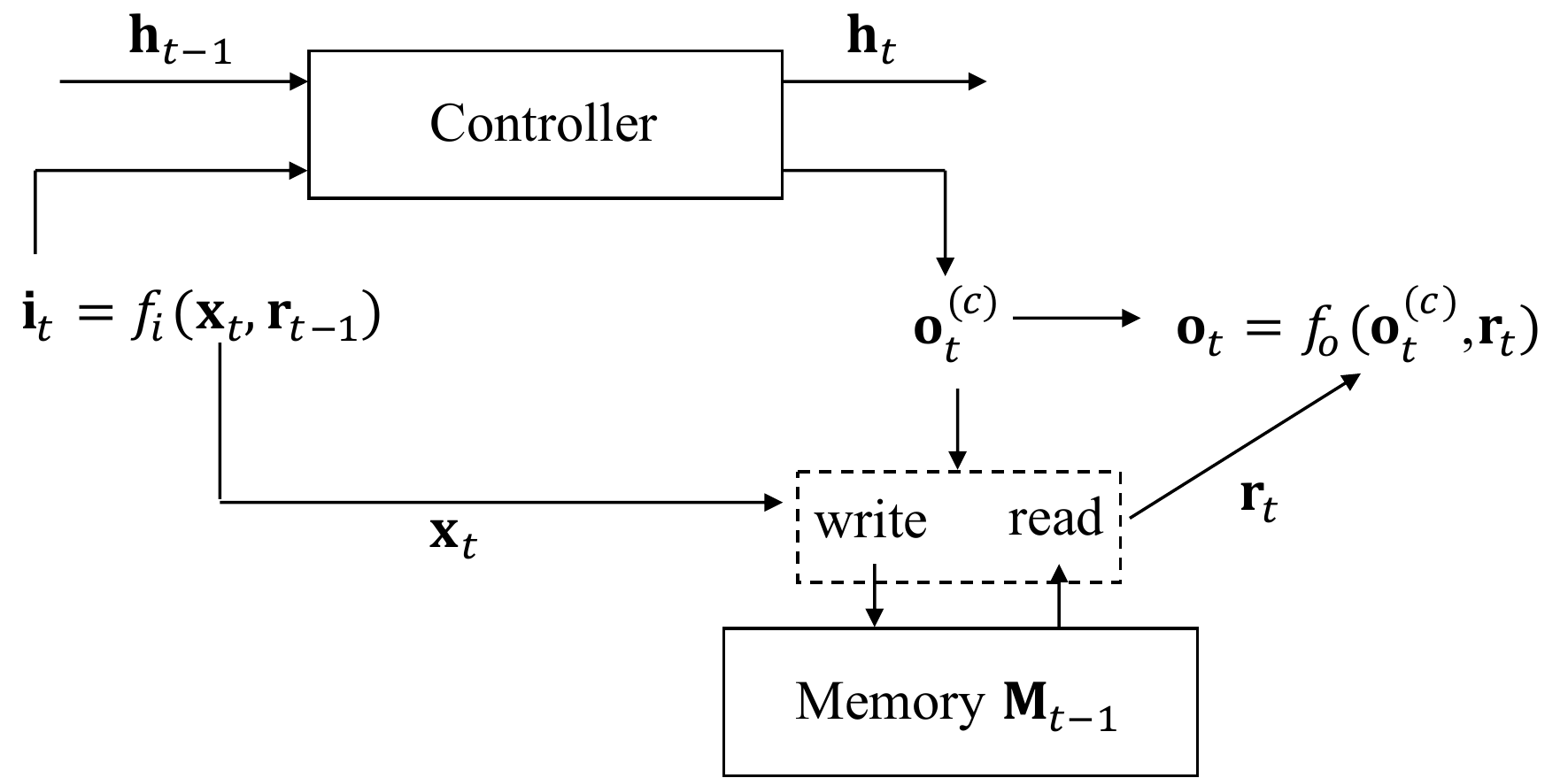}
  \caption{Proposed unified MANN framework}
  \label{f5}
\end{figure}

In order to compare different types of memory modules, we propose a unified framework for MANNs by fixing the controller and memory access interface. By abstracting the processing components used in stack augmented neural network~\citep{joulin2015inferring, yogatama2018memory} and neural Turing machines~\citep{graves2014neural}, the framework shown in Figure~\ref{f5} contains an LSTM controller, a memory module at time $t$ with $N$ cells represented as $\mathbf{M}_{t}=[\mathbf m_0\ \mathbf m_1\ \cdots\ \mathbf m_{N-1}]^T\in \mathbb{R}^{N\times d}$ equipped with a specific read-write method, such as push and pop actions of the stack memory extension.

The input $\mathbf i_t$ at time step $t$ to the controller is a combination of the system input $\mathbf x_t$ at time step $t$ and readout $\mathbf r_{t-1}$ at the last time step. Formally,
\begin{sequation}
\label{eq:input}
\mathbf i_t = f_i(\mathbf x_t, \mathbf r_{t-1}),
\end{sequation}where $f_i$ is a learnable function and we here take it simply to be a concatenation operation. The state of the LSTM controller is represented as $\mathbf h_t$, updated by the standard LSTM model:
\begin{sequation}
\label{eq:controller_state}
\mathbf h_t = \mathrm{LSTM}(\mathbf i_t, \mathbf h_{t-1}).
\end{sequation}

The controller output $\mathbf o_t^{(c)}=\mathbf h_t$ and the input $\mathbf{x}_t$ are then inputted to the write module and the read module, indicating how to interact with the memory. Finally, the readout $\mathbf r_t$ is combined with the controller output $\mathbf o_t^{(c)}$ to form the system output $\mathbf o_t$ at time step $t$:
\begin{sequation}
\label{eq:output}
\mathbf o_t = f_o(\mathbf o_t^{(c)}, \mathbf r_t),
\end{sequation}where $f_o$ is also a concatenation function.

In this paper, we experiment with two typical MANNs whose external memories are a \textbf{stack memory} and a \textbf{tape memory} respectively. To implement the two models under our proposed framework, we only need to specify the detailed read and write methods, which are shown in Section~\ref{section:stack} and~\ref{section:ntm} respectively. As a special case, an LSTM model is an instance of the framework without memory\footnote{The LSTM model contains internal memory actually. The term memory mentioned in this paper all refers to external memory, if without specific explanation.}.

\subsection{Stack memory}
\label{section:stack}
We adopt the stack-augmented neural network (SANN) proposed in~\cite{yogatama2018memory}. The readout at each time step is just the top of the stack:
\begin{sequation}\label{eq:stack_readout}
\mathbf r_t = \mathbf{M}_t[0].
\end{sequation}

For each write step, there are $2K+2$ possible actions $a_i$'s to choose:
\begin{itemize}
  \item $\mathrm{PUSH_k}$: Push the transformed current controller state $f_s(\mathbf h_t)$ onto the stack after $k\ (k=0,1,2,\cdots K)$ pops.
  \item $\mathrm{STAY_k}$: Keep the stack unchanged after $k\ (k=0,1,2,\cdots K)$ pops.
\end{itemize}Where $f_s$ is a learnable function and is implemented as a linear transformation.

In order to ensure the model differentiable, the write step for the stack is formalized as the expectation of the memory after one of the $2K+2$ specific actions $a_i$:
\begin{sequation}
\label{eq:stack_update}
  \mathbf{M} := \sum_{i}p(a_i|\mathbf{M})\cdot \mathbf{M}_{a_i},
\end{sequation}where the $\mathbf{M}_{a_i}$ is the memory after the action $a_i$ is adopted, and the probabilities of the write actions $\mathbf{a}=[a_0\ a_1\ \cdots\ a_{2K+1}]^T$ is computed using the current memory $\mathbf{M}_t$ and the system input $\mathbf{x}_t$\footnote{Although the policy is parameterized in a recursive formula in the original paper, we find this simple setting is powerful enough.}:
\begin{sequation}
\label{eq:stack_policy}
p(\mathbf{a}|\mathbf{M}_t, \mathbf{x}_t) = \mathrm{softmax}(\mathbf{W}_{a\mathbf{M}}\mathrm{Conv1d}(\mathbf{M}_t)+\mathbf{W}_{ax}\mathbf x_t+\mathbf b_{a}),
\end{sequation}where the $\mathrm{Conv1d}$ is a two-channel $1$-D convolutional operation with a size-$2$ kernel over the $N$ memory cells.

\subsection{Tape memory}
\label{section:ntm}
As the memory in a standard Turing machine is called a tape, we here name the class of neural Turing machines by TANNs, i.e. tape augmented neural networks. The tape memory module is adopted as mentioned in~\cite{graves2014neural}, with random access read-write steps totally controlled by the controller.

Both the read and write actions contain two preparation steps before the actual interaction with the memory: 1) an analysis step to the controller state and 2) an addressing step. The detailed formulas of these two steps are omitted here. After the addresses to read and write are determined, the readout is the expectation of the memory cells over distribution $\mathbf w^r$:
\begin{sequation}
\label{eq:ntm_readout}
\mathbf r_t = \sum_{i=0}^{N-1}\mathbf w_t^r[i]\mathbf{M}[i],
\end{sequation}where $[i]$ means to index the $i$th row of the vector or matrix. And the write step:
\begin{sequation}
\label{eq:tape_update}
\mathbf{M}_t = \mathbf{M}_{t-1}\odot (1-\mathbf w_t^w \mathbf e_t^T) + \mathbf w_t^w\mathbf a_t^T,
\end{sequation}which is combining the influence of erase vector $\mathbf e_t$ and add vector $\mathbf a_t$ on the memory $\mathbf{M}_{t-1}$ over the distribution $\mathbf w^w$.

\section{Experiment}
In the experiment, we want to figure out which kind of and how memory module helps generalize on two algorithm tasks including reversing a random sequence (called mirror task) and evaluating arithmetic expressions (called \emph{M10AE} task). Simple RNN (SimpRNN) and LSTM are adopted as two baseline models, which represent neural networks without external memory modules.

\begin{table}[H]
\begin{tabular}{lccc}
\toprule
task & input & output & measure\\
\midrule
mirror & $\mathbf{x}_1\mathbf{x}_2\cdots \mathbf{x}_{L}$ & $\mathbf{x}_{L} \mathbf{x}_{L-1}\cdots \mathbf{x}_1$ & input length\\
\emph{M10AE} & $8+6*3/2-4$ & $4$ & \#LPO \\ 
\bottomrule
\end{tabular}
\caption{Task examples. The $\mathbf{x}_i$ represents the $i$th random binary vector. \#LPO is short for number of low-priority operators. Note that for \emph{M10AE} the \emph{/} represents a modulo operator and each intermediate result is followed by a modulo-$10$ operation. For an instance, the evaluation procedure of the example in the table can be done in $4$ steps: i.$(6*3)\%10=8$, ii.$(8~\%~2)\%10=0$, iii.$(8+0)\%10=8$, iv.$(8-4)\%10=4$.}
\label{tb:examples}
\end{table}

\subsection{Experimental settings}
For mirror task, the input is a sequence of binary vectors whose size is $9$. During encoding (input) stage, the inputs are randomly sampled from Bernoulli distribution with $p=0.5$ and the inputs are zero vector during decoding (output) stage. This setting is similar to that of copy task in \cite{graves2014neural}. Both the number $N$ of and size $M$ of memory cells are $20$. The controller dimension is $100$. For \emph{M10AE} task, the input embeddings $\mathbf x_t$'s are trainable parameters from random initialization. The input embedding dimension $100$. The dimension of a memory cell and a controller state is set to be equal to the input embedding dimension. The number of memory cells is chosen from [$5$, $10$, $20$]. We adopt training using Adam algorithm with batch size chosen from [$32$, $64$, $128$, $256$] and the learning rate chosen from [$0.001$, $0.0001$]. The hyper-parameters are tuned on a development set.

\subsection{Mirror}
First, we are interested in exploring whether SANN and TANN are able to learn to output the input sequence in a reverse order, which we call mirror task. An example of this task is shown in the first row of Table~\ref{tb:examples}. We append a delimiter at the end of the input to tell the model when to output, and it is noted as $\langle\mathrm{EOS}\rangle$. We adopt the length of input sequence as the difficulty measure of each sample.
\subsubsection{Result}
\begin{figure}[H]
  \centering
  \begin{tabular}{cc}
  \begin{minipage}{0.45\linewidth}
      \centerline{\includegraphics[height=2.8cm]{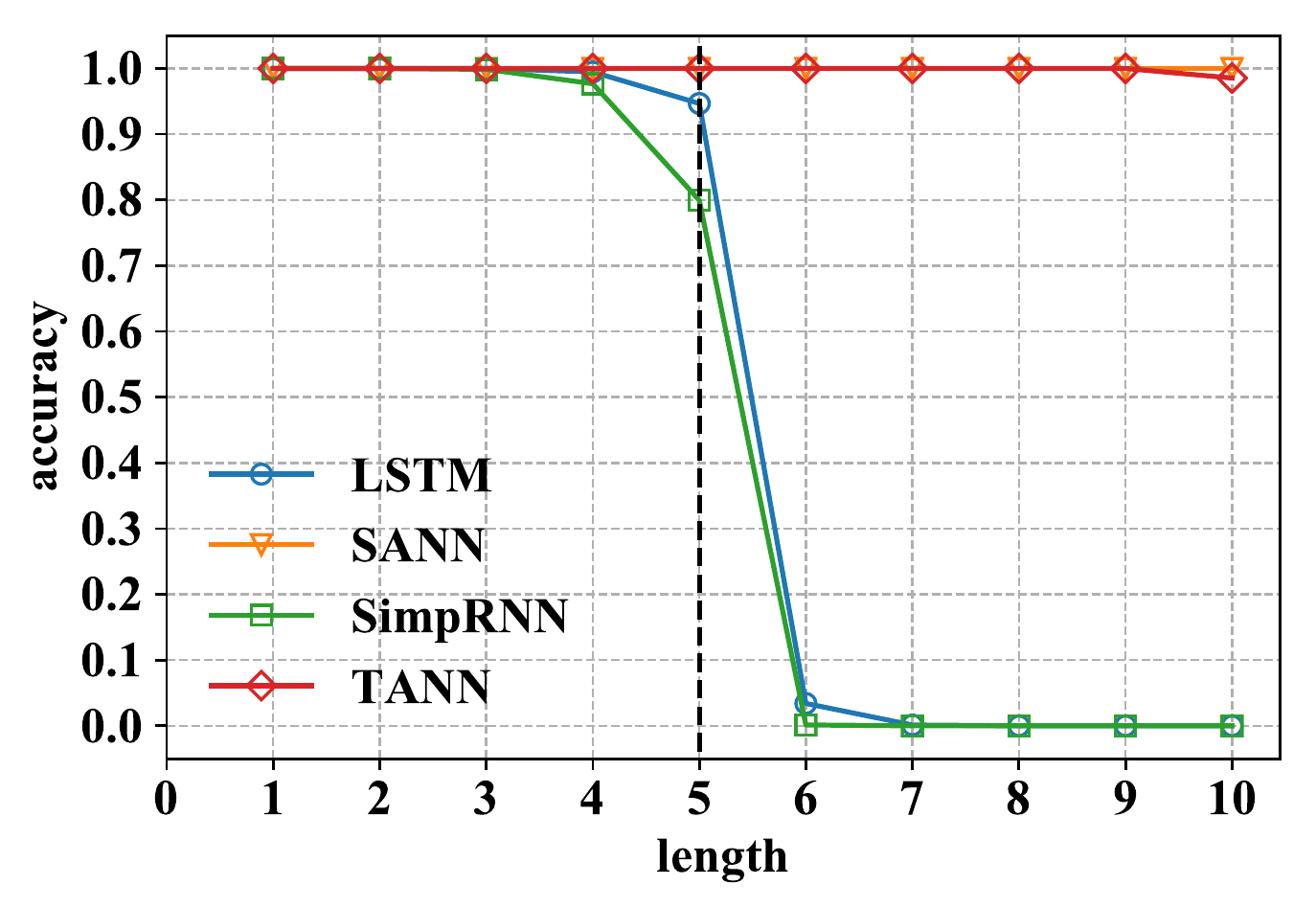}}
      \centerline{a. test performance}
  \end{minipage}
  &
  \begin{minipage}{0.45\linewidth}
      \centerline{\includegraphics[height=2.8cm]{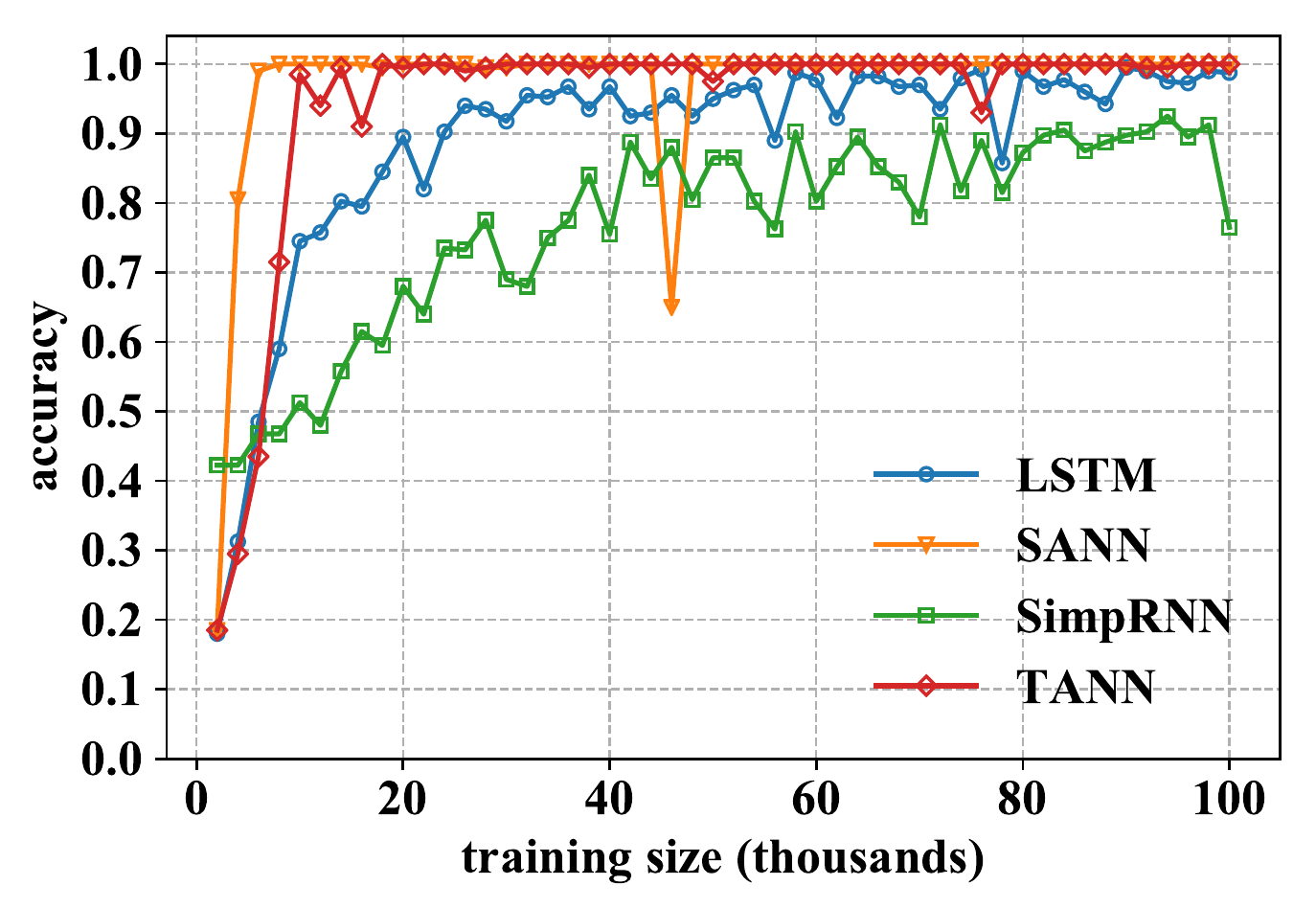}}
      \centerline{b. learning curves}
  \end{minipage}
  \end{tabular}

  \caption{(a) Test performance along with different input length for mirror task. Black dash line indicates the maximum length of input sequence during the training stage. (b) Learning curves for each model on mirror task in the training stage, whose y-axis indicates the performance on the training set. }
  \label{fig:mirror_len}
\end{figure}

The maximum length of input sequence is $5$ during the training stage, and the maximum length is extended to $10$ when testing. Here we view a prediction as correct only if the whole output sequence is the same as that of the input. The result is shown in Figure~\ref{fig:mirror_len}. We can find that both SANN and TANN can generalize beyond input length of training samples. Figure~\ref{fig:mirror_len}b shows that SANN converges faster than TANN, which corresponds with the intuition that stack memory is more suitable for this task. 
\subsubsection{Analysis}
\label{sec:analysis_mirror}
Since TANN and SANN can generalize greatly, we here analyze both of them to investigate what strategy they have induced. In order to gain a general averaged insight into what mechanism underlying these two models on mirror task, we generate $500$ samples with the same length, whose each input binary vector is restricted in the binary format of $1, 2, \cdots, 9$. These numbers can be viewed as the labels of the samples, which helps index the input vectors. And all the analysis for mirror task is based on the $500$ samples.
\begin{figure}[H]
  \centering
  \begin{tabular}{cc}
  \begin{minipage}{0.45\linewidth}
      \centerline{\includegraphics[height=2.8cm]{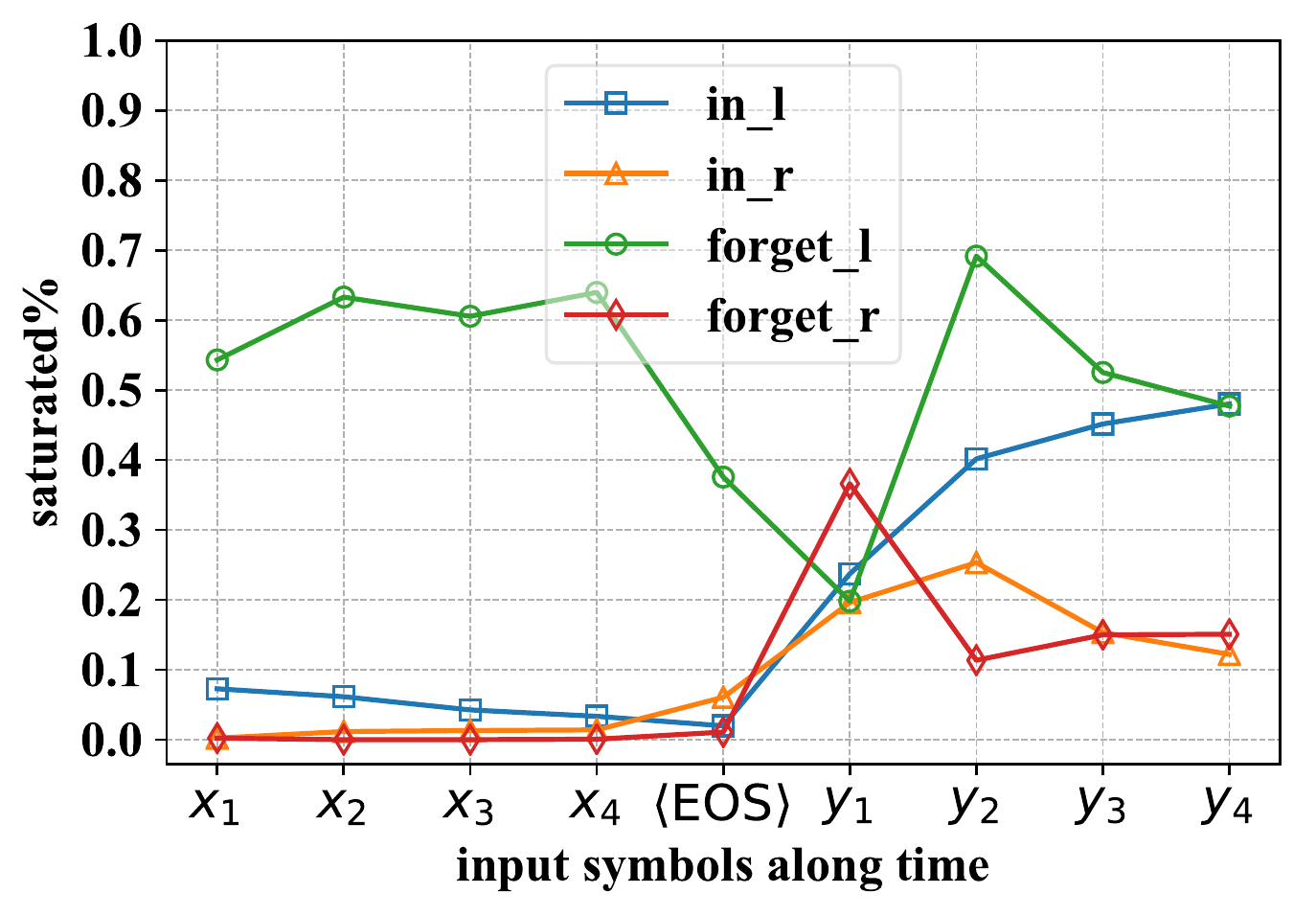}}
      \centerline{a. controller gate (TANN)}
  \end{minipage}
  &
  \begin{minipage}{0.45\linewidth}
      \centerline{\includegraphics[height=2.8cm]{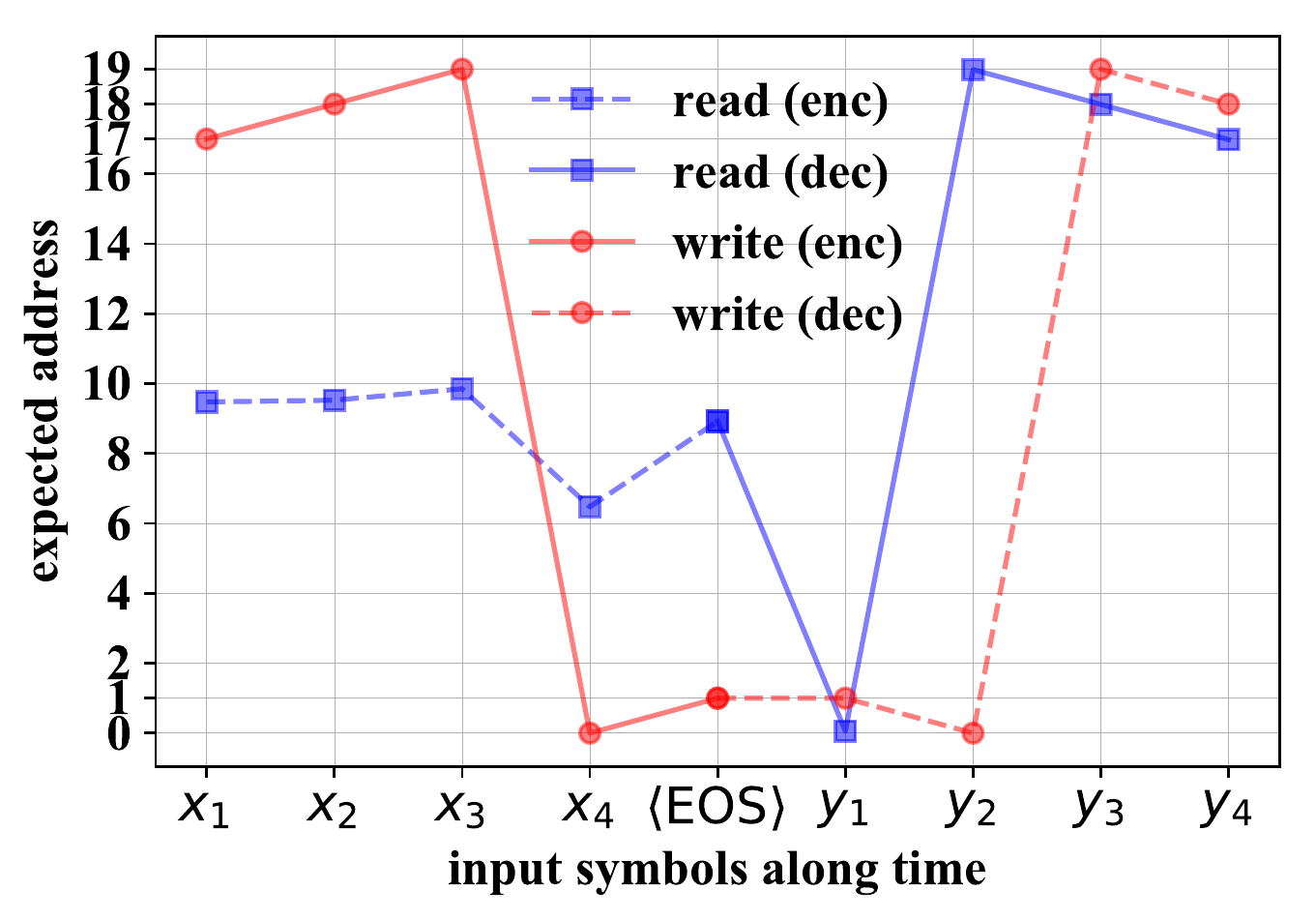}}
      \centerline{b. read-write policy (TANN)}
  \end{minipage}\\
  \begin{minipage}{0.45\linewidth}
      \centerline{\includegraphics[height=2.8cm]{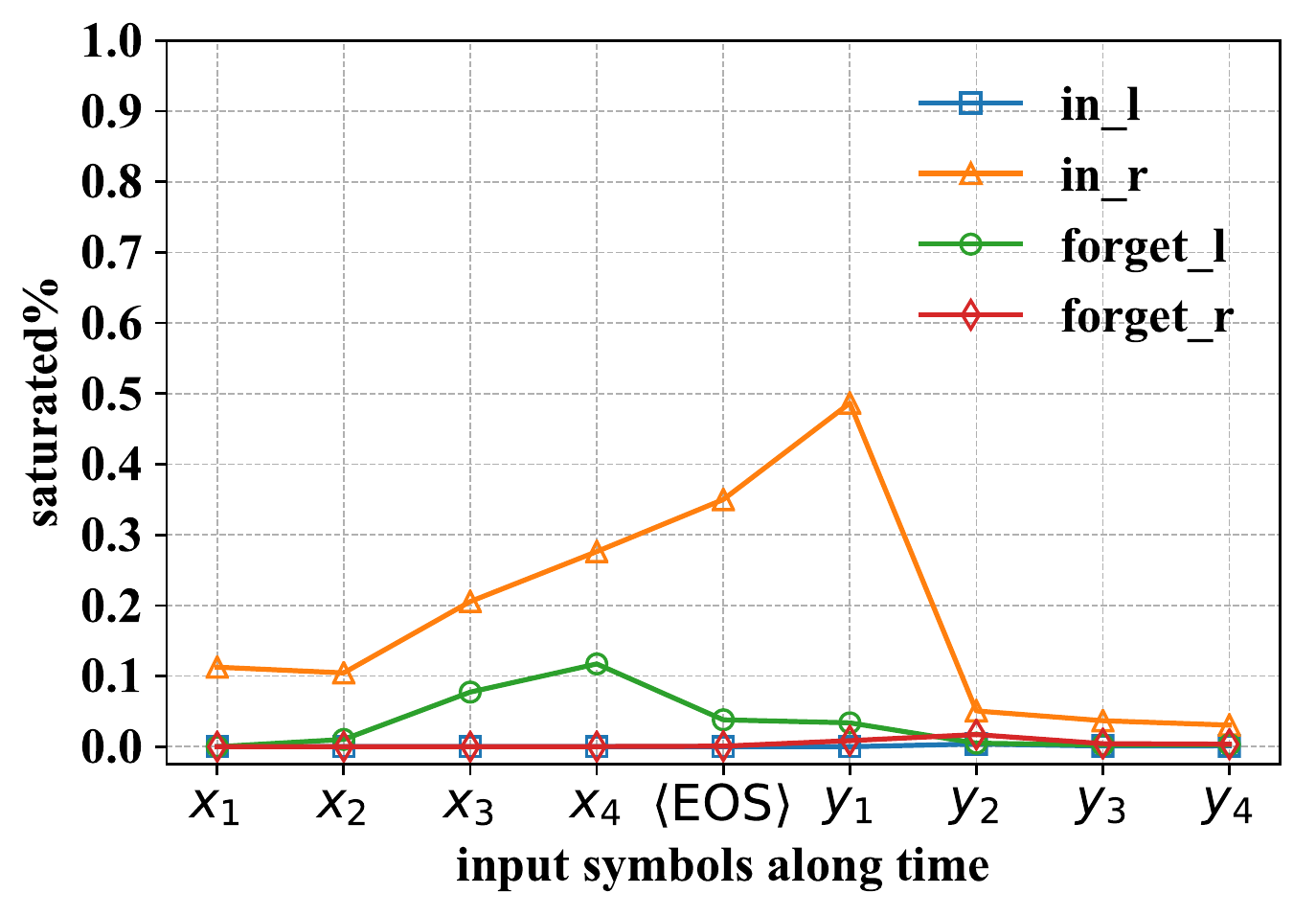}}
      \centerline{c. controller gate (SANN)}
  \end{minipage}
  &
  \begin{minipage}{0.45\linewidth}
      \centerline{\includegraphics[height=2.8cm]{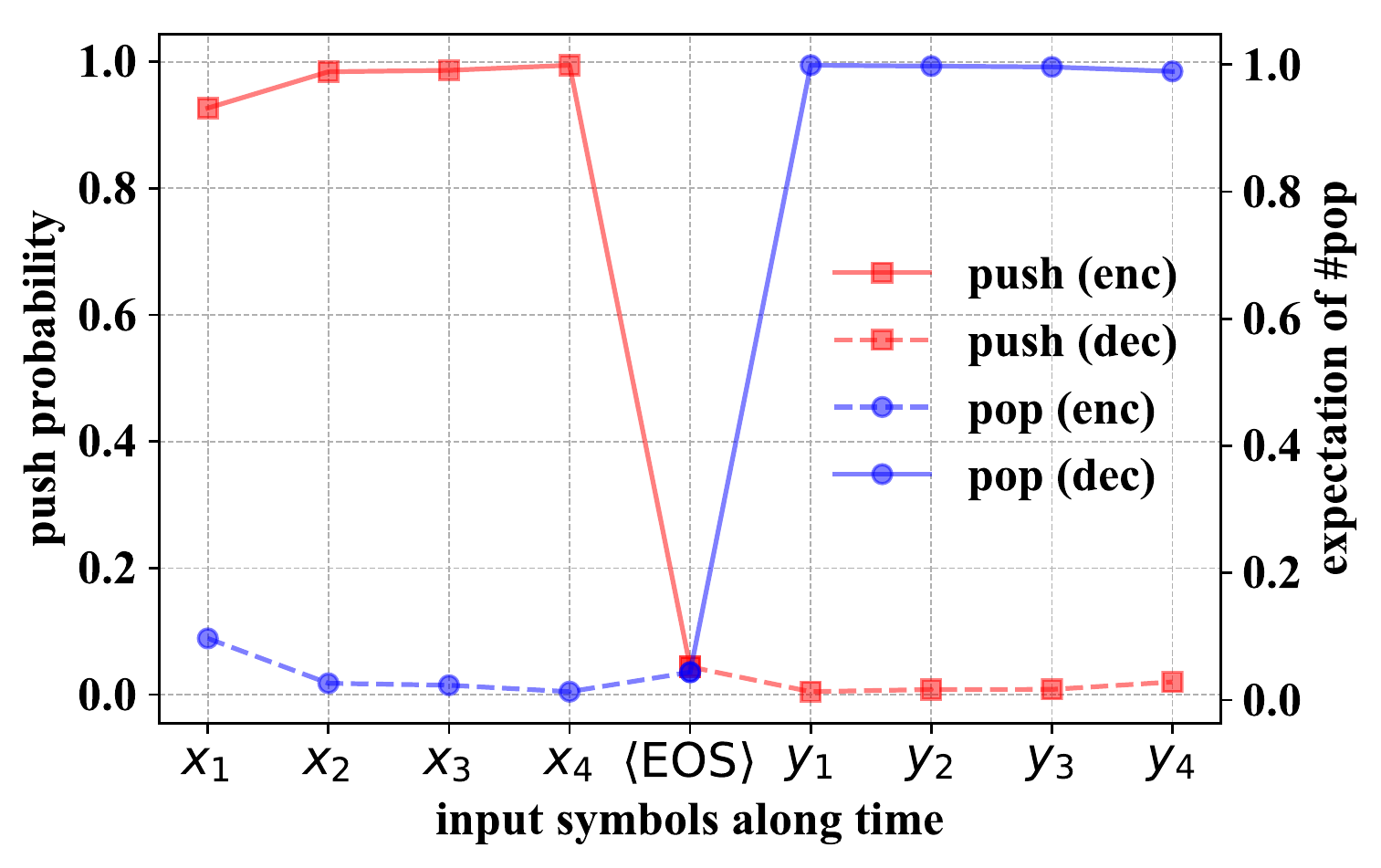}}
      \centerline{d. push-pop policy (SANN)}
  \end{minipage}
  \end{tabular}
  \caption{Averaged visualization about (a and c) controller gate and (b and d) read-write policy for TANN and SANN on mirror task. Note that all the plots are derived from being averaging over $500$ random samples. The x-axis shows each time step represented by input $\mathbf{x}_i$ or output $\mathbf{y}_i$. The \emph{$\langle\mathrm{EOS}\rangle$} represent the input delimiter.}
  \label{fig:mirror_visual}
\end{figure}
We first are interested in investigating how the controller gates change on mirror task. Specifically, we plot in Figure~\ref{fig:mirror_visual}a and~\ref{fig:mirror_visual}c the averaged saturation ratio~\citep{DBLP:journals/corr/KarpathyJL15} of the input gates and forget gates of the controller along with each input. Here a gate is defined right-saturated if its value is larger than $0.9$ and defined left-saturated if its value is less than $0.1$. Comparing these two figures, we can find that both TANN and SANN are sensitive of the delimiter in terms of each controller gates, after which dramatic changes of saturation rate appear. The change of controller gates of TANN seem more complicated than that of SANN, which indicates it is much easier to control a stack memory to finish the mirror task than a tape memory. The early convergence of SANN on mirror task can also support this idea.

We then visualize the read-write and push-pop policies for TANN and SANN respectively. For read-write policy of TANN, we average the expected address over the $500$ samples. The expected address $p_t$ for read and write operations at time step $t$ can be calculated as:
\begin{sequation}\label{eq:expected_address}
p_t = \sum_{i=0}^{N-1}\mathbf{w}_t[i]\cdot i,
\end{sequation}
where the $\mathbf{w}_t$ is either $\mathbf{w}_t^r$ or $\mathbf{w}_t^w$ to get the expected address for read or write. For SANN, the push probability is just the probability sum mass of all type of push actions:
\begin{sequation}\label{eq:push_prob}
P(\mathrm{PUSH}|\mathbf{M}_t) = \sum_{k=0}^{K-1}P(\mathrm{PUSH}_k|\mathbf{M}_t),
\end{sequation}
and the expected number of times to pop (noted as $n_{\mathrm{pop}}$) at time step $t$ can be calculated as:
\begin{sequation}\label{eq:stack_policy_visual}
n_{\mathrm{pop}} = \sum_{k=0}^{K-1}P(\mathrm{POP}_k|\mathbf{M}_t)\cdot k.
\end{sequation}
The result is shown in Figure~\ref{fig:mirror_visual}b and~\ref{fig:mirror_visual}d. We can find that both for TANN and SANN the policies for encoding (solid red lines) and decoding (solid blue lines) information are opposite to each other, indicating reversing in their own ways.

\begin{figure}[H]
  \centering
  \begin{tabular}{ccc}
  \begin{minipage}{0.30\linewidth}
      \centerline{\includegraphics[height=2.6cm]{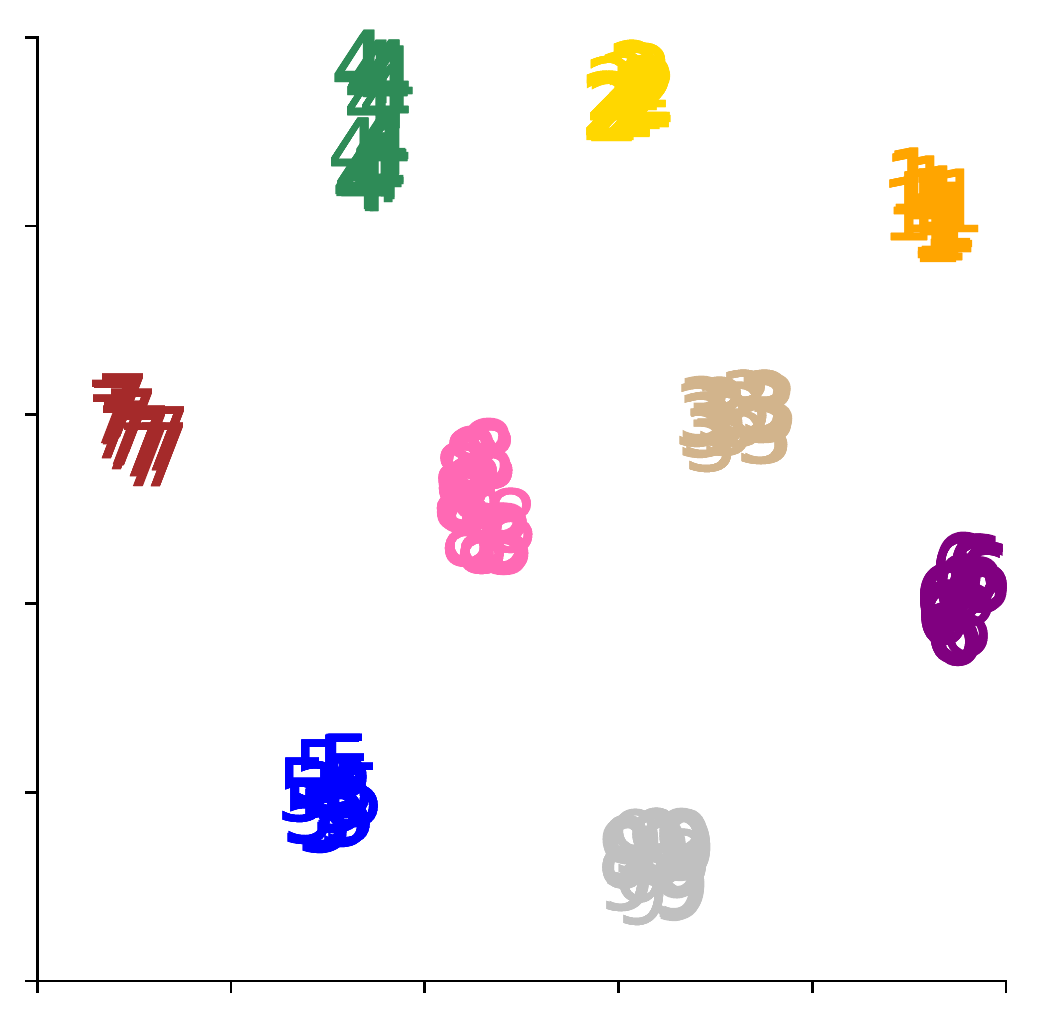}}
      \centerline{a. ($1$, $18$); $\mathbf{x}_1$}
  \end{minipage}
  &
  \begin{minipage}{0.30\linewidth}
      \centerline{\includegraphics[height=2.6cm]{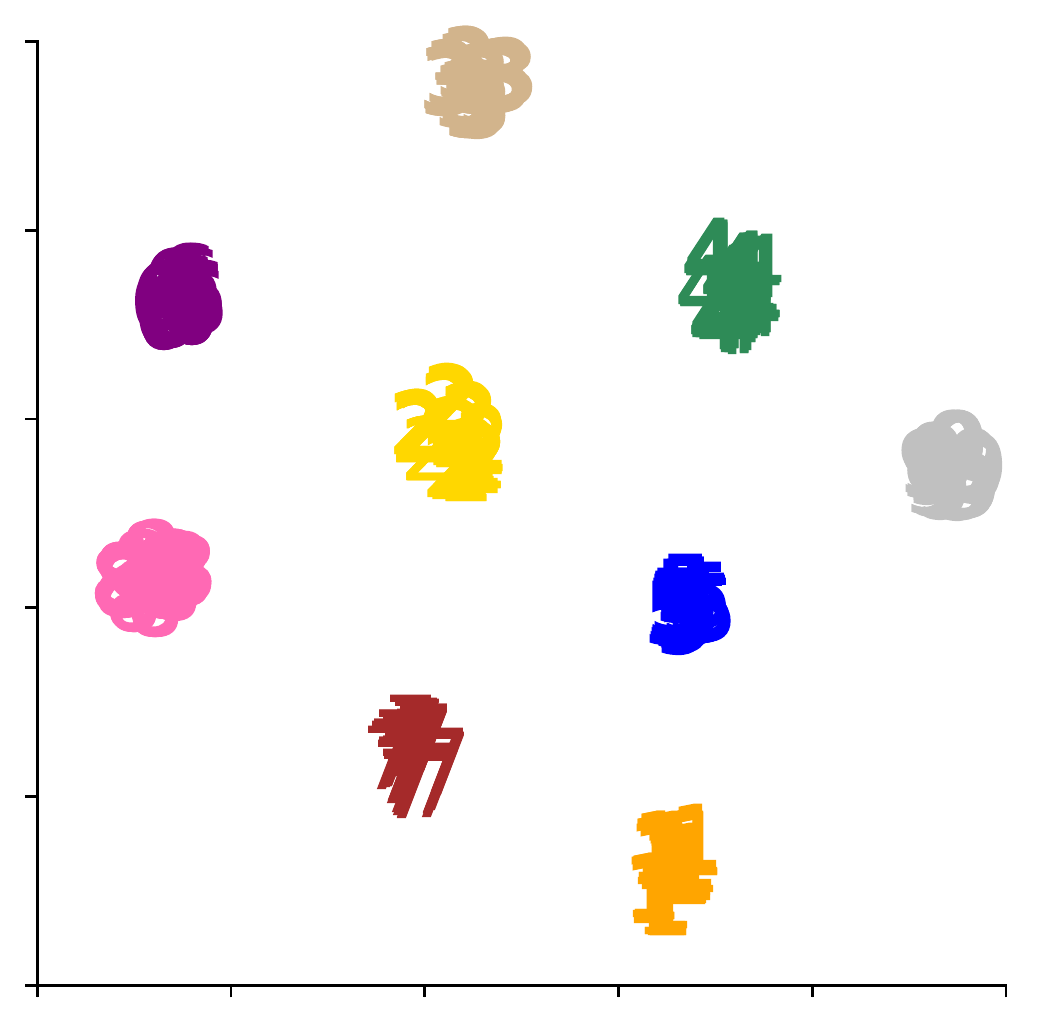}}
      \centerline{b. ($2$, $19$); $\mathbf{x}_2$}
  \end{minipage}
  &
  \begin{minipage}{0.30\linewidth}
      \centerline{\includegraphics[height=2.6cm]{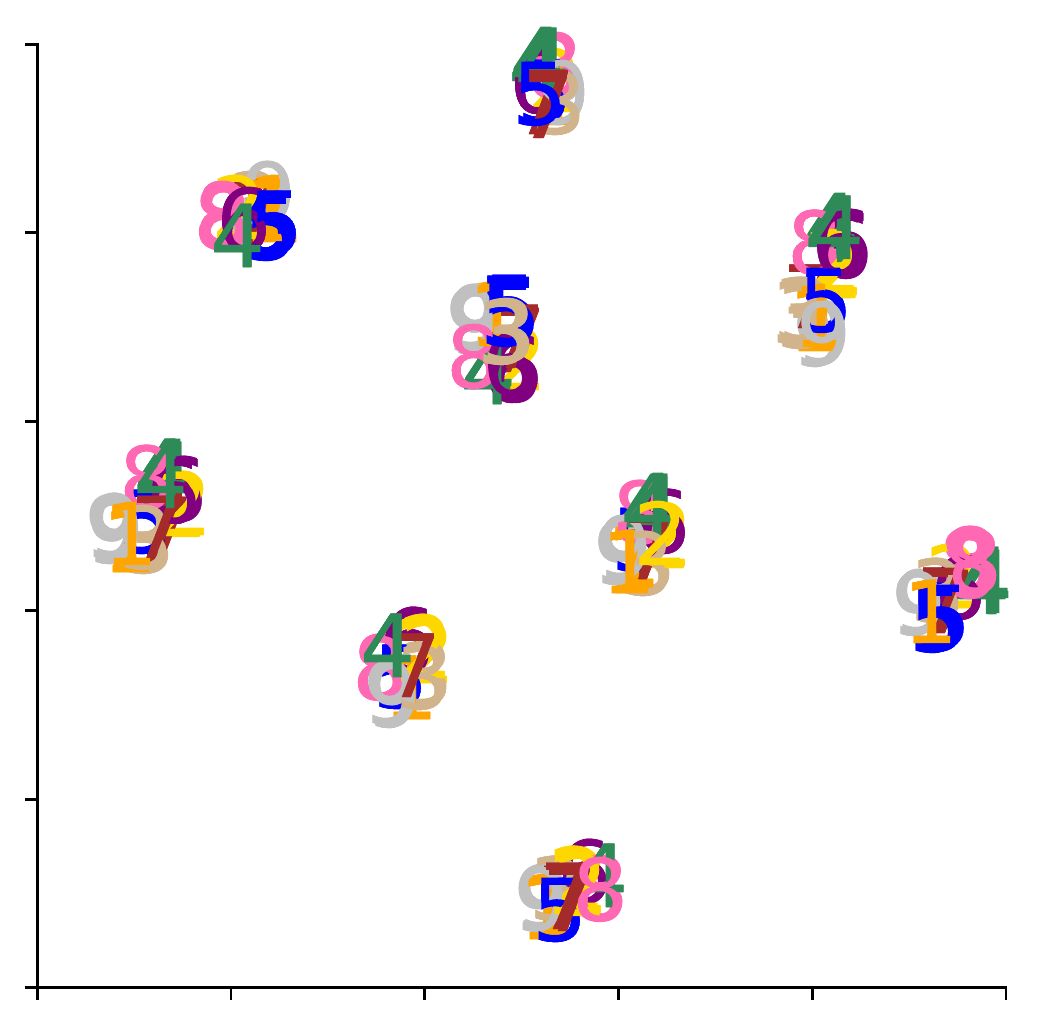}}
      \centerline{c. ($1$, $18$); $\mathbf{x}_0$}
  \end{minipage}\\
  \begin{minipage}{0.30\linewidth}
      \centerline{\includegraphics[height=2.6cm]{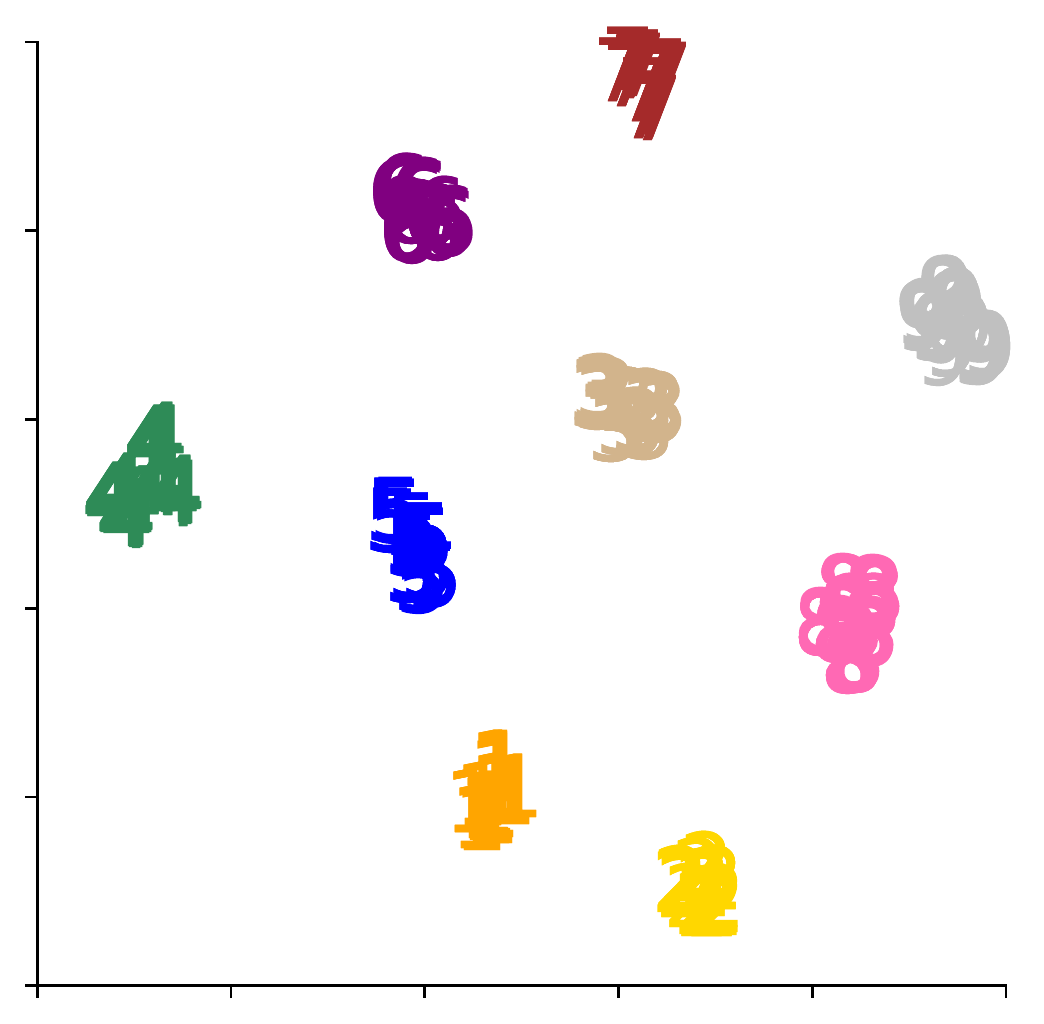}}
      \centerline{d. ($1$, $0$); $\mathbf{x}_1$}
  \end{minipage}
  &
  \begin{minipage}{0.30\linewidth}
      \centerline{\includegraphics[height=2.6cm]{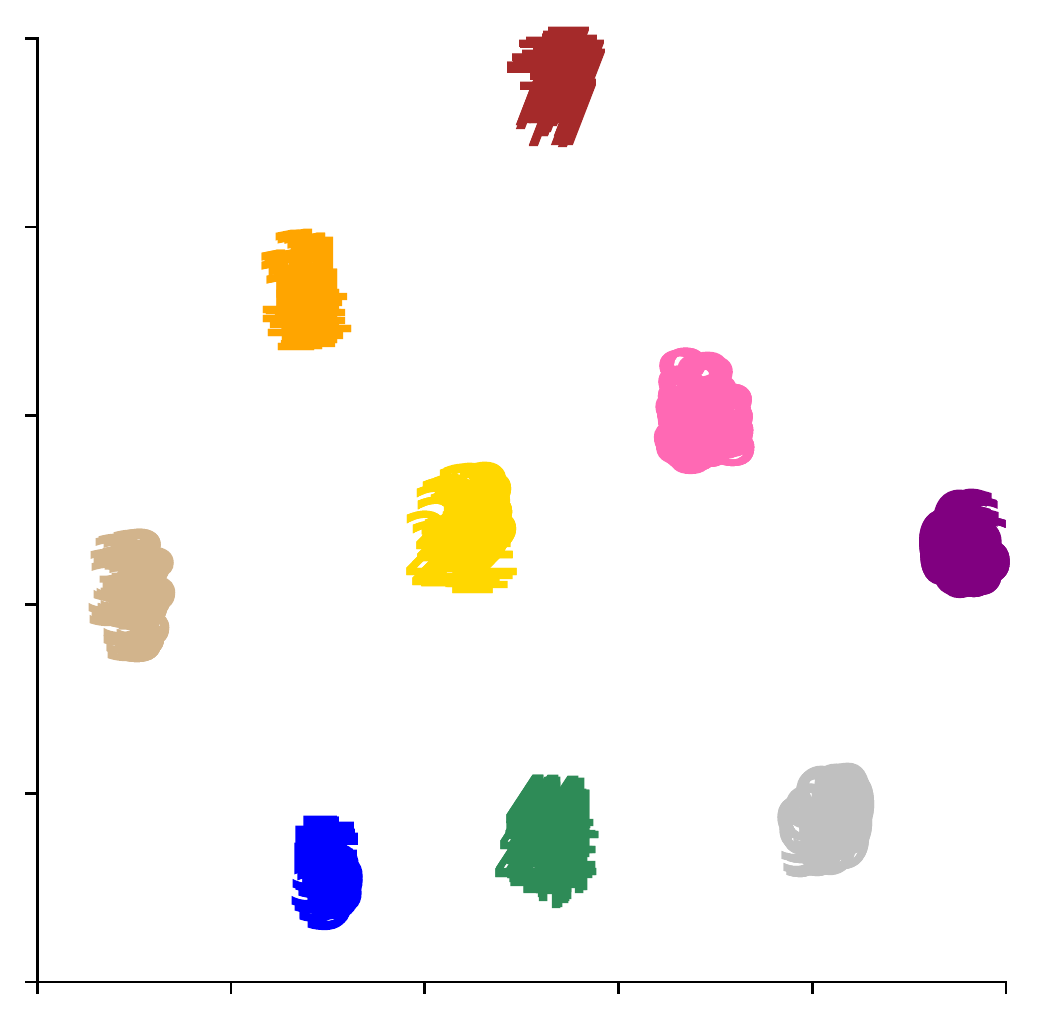}}
      \centerline{e. ($2$, $0$); $\mathbf{x}_2$}
  \end{minipage}
  &
  \begin{minipage}{0.30\linewidth}
      \centerline{\includegraphics[height=2.6cm]{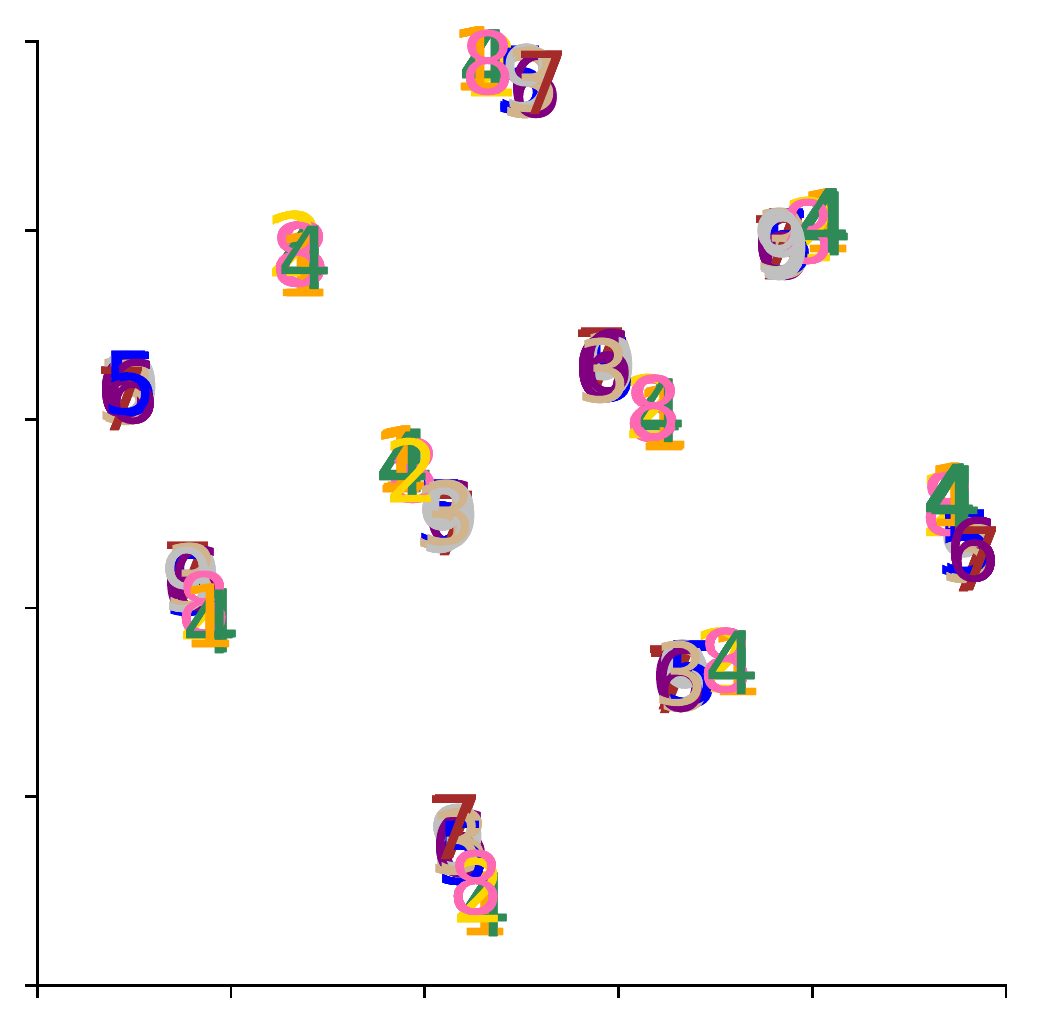}}
      \centerline{f. ($1$, $1$); $\mathbf{x}_1$}
  \end{minipage}\\
  
  \end{tabular}
  \caption{Six examples of qualitative evaluation result of hypothesis information encoded in memory cells of TANN (a, b and c) and SANN (d, e and f). The parts before the semicolon in the captions are the time step and the memory cell address, noted in the form of ($\left\langle {time\ step} \right\rangle$, $\left\langle {cell\ address} \right\rangle$). And after the semicolon is the hypothesis semantics of the memory cells, represented as certain position of the input sequence.}
  \label{fig:mirror_verification}
\end{figure}

Based on these findings above, we make hypothesis about what strategy learned by TANN by the following pseudocode:
\begin{algorithmic}[1]
\Require 
  sequence of binary vectors $s=\mathbf{x}_1, \mathbf{x}_2, \cdots, \mathbf{x}_L$
\Ensure
  the reverse of the input sequence
  \State $\mathbf{M}\in \mathbb R^{N\times M}$ \Comment{initialize the memory randomly}
  \State $w^w \gets 17$ \Comment{initialize the writing-to address}
  
  \For{ each input $\mathbf{x}_t$ in $s$} \Comment{encoding stage}
    \If{$\mathbf{x}_t$ is a delimiter} \Comment{the end of the input}
      \State $w^r \gets (17+L-1)\%N$ \Comment{initialize the reading-from address}
    \Else 
      \State $\mathbf{M}[w^w] \gets \mathbf{x}_t$ \Comment{write to store $\mathbf{x}_t$}
      \State $w^w \gets (w^w + 1) \% N$ \Comment{point to the next cell}
    \EndIf
  \EndFor

  \For {$t$ in $1, 2, ..., L$} \Comment{decoding stage}
    \State output {$\mathbf{M}[w^r]$} \Comment{read and output}
    \State $w^r \gets (w^r - 1) \% N$ \Comment{point to the next cell}
  \EndFor
\end{algorithmic}
And the hypothesis strategy for SANN on mirror task is:
\begin{algorithmic}[1]
\Require 
  sequence of binary vectors $s=\mathbf{x}_1, \mathbf{x}_2, \cdots, \mathbf{x}_L$
\Ensure
  the reverse of the input sequence
  \State empty the stack $S$
  
  \For{ each input $\mathbf{x}_t$ in $s$} \Comment{encoding stage}
    \If{$\mathbf{x}_t$ is not a delimiter}
      \State push the $\mathbf{x}_t$ to $S$ \Comment{store information}
    \EndIf
  \EndFor

  \For {$t$ in $1, 2, ..., L$} \Comment{decoding stage}
    \State output the top of $S$ \Comment{read and output}
    \State pop the top of $S$
  \EndFor
\end{algorithmic}
The next goal is to verify these hypotheses above. To this end, we evaluate the hypothesis information encoded in each memory cell by our proposed qualitative verification method. This method is based on the assumption that if the intermediate results are the same in a certain step of the hypothesized strategy, then their distributed representations in the memory should be similar as well. In detail, this includes $4$ steps:
\begin{enumerate}
  \item Collecting the memory cells $[\mathbf{m}_t^{(1)}, \mathbf{m}_t^{(2)}, \cdots, \mathbf{m}_t^{(l)}]$ at the same position in the memory at the same time step.
  \item Labelling the memory cells with the corresponding results $[\hat{y}_t^{(1)}, \hat{y}_t^{(2)}, \cdots, \hat{y}_t^{(l)}]$ derived from the candidate strategy to get pairs $\left\{ (\mathbf{m}_t^{(i)}, \hat{y}_t^{(i)})\right\}_{i=1}^{l}$.
  \item Using t-SNE~\citep{maaten2008visualizing} to visualize the labelled memory cell vectors.
  \item If there appear the clear labelled clusters, then the candidate strategy is reasonable.
\end{enumerate}
The examples are shown in Figure~\ref{fig:mirror_verification}. The compact labelled clusters in Figure~\ref{fig:mirror_verification}a,~\ref{fig:mirror_verification}b (for TANN) and Figure~\ref{fig:mirror_verification}d,~\ref{fig:mirror_verification}e (for SANN) support the hypothesis semantics shown in the captions of each subfigure. We also include negative examples in Figure~\ref{fig:mirror_verification}c and~\ref{fig:mirror_verification}f to exemplify when the evaluation result shows the hypothesis is not reasonable. Since the input sequence are randomly sampled, labelling guided by a wrong hypothesis can not cover all the samples and this mismatch will also show up in the visualization as chaotic labels.

\subsection{M10AE}
After analysis on mirror tasks, we are then interested in extending the interpretation procedure to the harder case. We propose a new simple algorithm task named \emph{modulo-$10$ arithmetic expressions}, \emph{M10AE}, in which each input sequence is in the form of an arithmetic expression without parentheses and the output is the evaluation result of it. In order to make the task tractable to analyze while preserving the recursive nature inside it, we add the following constraints:
\begin{itemize}
\item Each numeral is limited in $1, 2, \cdots, 9$.
\item The \emph{/} represent a modulo operator.
\item Each intermediate result is modulo by $10$.
\item The \emph{*}, \emph{/} have higher priority than \emph{+} and \emph{-}.
\end{itemize}
An example is shown in the second row of Table~\ref{tb:examples}. Based on the constraints, the evaluation results can only range in integers from $0$ to $9$, and thus we formalize the task as a $10$-class classification. Since the computation procedure interrupts when encountering low-priority operators and thus larger number of low-priority operators (\#LPO) means larger memory burden, we take \#LPO to be the difficulty measure for \emph{M10AE}.

Recently, \cite{hupkes2018visualisation} and \cite{JacobLSB18} have proposed two similar tasks. However, they formalize the task as either a regression problem or a two-digit sequence prediction problem, without consideration of the intermediate results. These settings are poor at restricting the space of intermediate results, whose distribution is much sparser, and this hinders us from verifying a candidate strategy empirically. By contrast, in \emph{M10AE}, the result is given as a classification label, and both the intermediate results and the final results range in integers from $0$ to $9$ due to modulo-$10$ design. This leads to abundant samples for each intermediate category.

\subsubsection{Result}
\begin{figure}[H]
  \centering
  \begin{tabular}{cc}
  \begin{minipage}{0.45\linewidth}
      \centerline{\includegraphics[height=2.8cm]{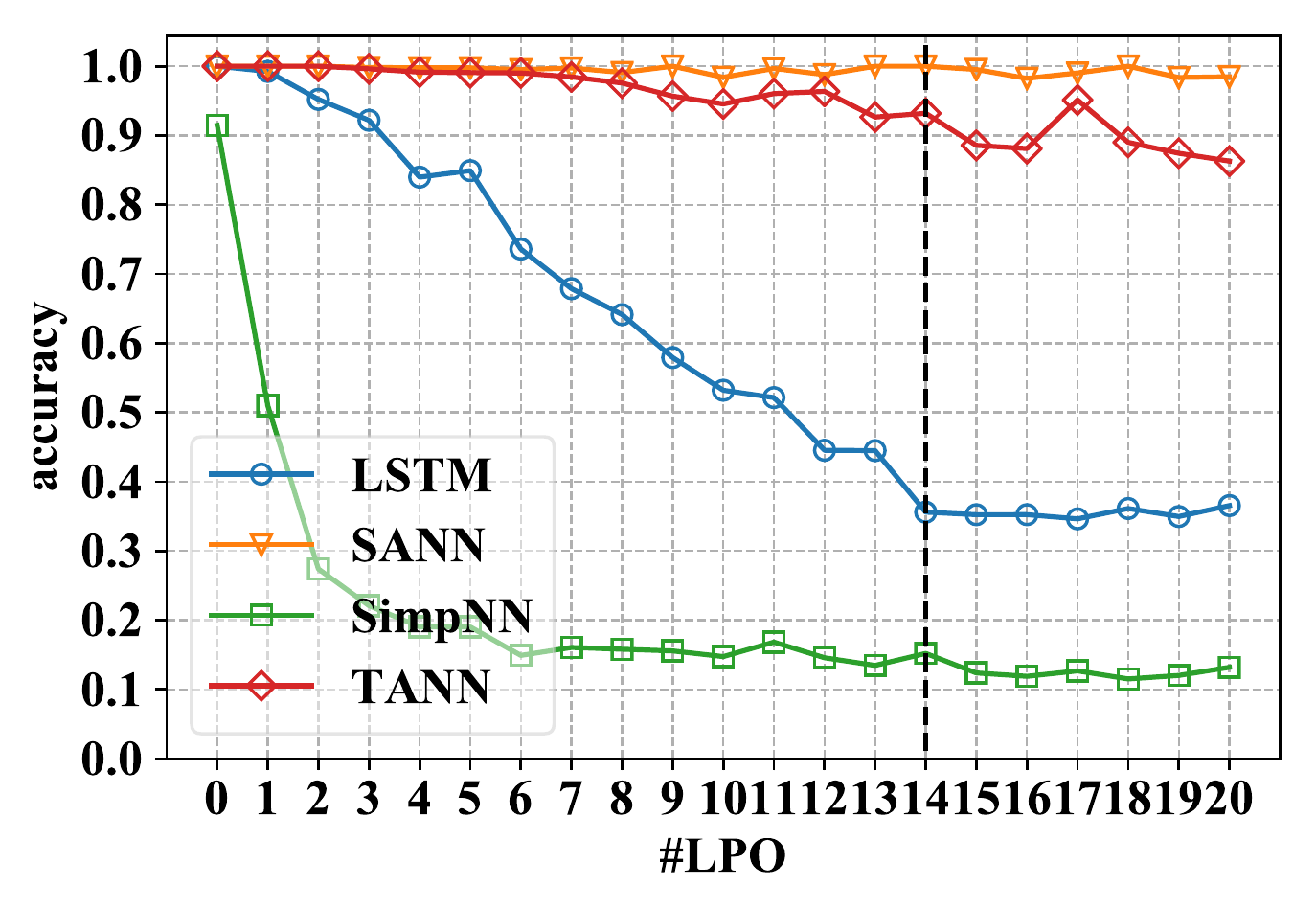}}
      \centerline{a. test performance}
  \end{minipage}
  &
  \begin{minipage}{0.45\linewidth}
      \centerline{\includegraphics[height=2.8cm]{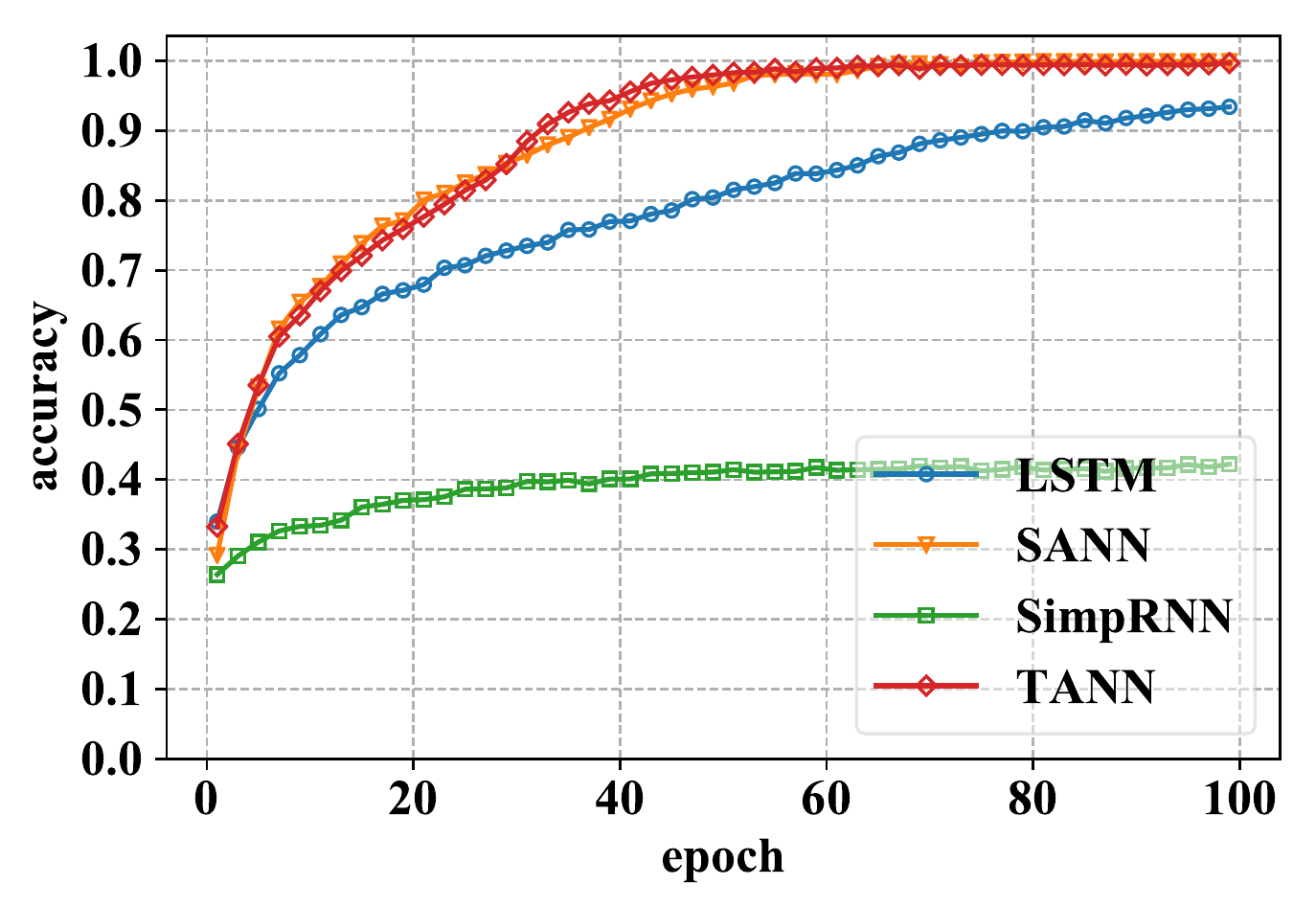}}
      \centerline{b. learning curves}
  \end{minipage}
  \end{tabular}

  \caption{(a) Test performance along with different input length for \emph{M10AE} task. Black dash line indicates the maximum \#LPO of input sequence during the training stage. (b) Learning curves for each model on \emph{M10AE} task in the training stage.}
  \label{fig:m10ae_nlpo}
\end{figure}

$80$ and $20$ thousand examples are generated for training and valiation, respectively. The maximum \#LPO is $14$ for training and $20$ for validation. The results are shown in Figure~\ref{fig:m10ae_nlpo}. Overall, the performance of all the models except SANN drops quickly with the increase of \#LPO. Figure~\ref{fig:m10ae_nlpo}a indicates the SANN has learned to generalize on this task. In addition, the models with an external memory are better than the ones without any external memory (i.e. LSTM and SimpRNN). By contrast, the baseline model simpRNN performs extremely bad, indicating that the internal memory of LSTM is crutial for this task. As shown in Figure~\ref{fig:m10ae_nlpo}b, all the models converge slowly, which indicates the complexity of the task.

\subsubsection{Analysis}
As the SANN seems to induce a stable strategy, we then analyze it. We generate $500$ samples with the same structure to get general averaged patterns. And the analysis is all based on these $500$ examples. These examples are noted as $\langle N_0 \rangle \langle L_0 \rangle \langle {N_1} \rangle \langle {H_1} \rangle \langle {N_2} \rangle \langle {H_2} \rangle \langle {N_3} \rangle \langle {L_1} \rangle \langle {N_4} \rangle$, where $\langle {N_i} \rangle$ represents the $i$th numeral, $\langle {L_i} \rangle$ represents the $i$th low-priority operator (i.e. \emph{+} or \emph{-}), and $\langle {H_i} \rangle$ represents the $i$th high-priority operator (i.e. \emph{*} or \emph{/}). One samples is \emph{8+6*3/2-4}. Other samples generated by substituting the operators and numbers at the same position into the other symbols from the same category, e.g., \emph{*} to \emph{/}, \emph{+} to \emph{-}, \emph{1} to \emph{2}, and etc.. 

\begin{figure}[H]
  \centering
  \begin{tabular}{cc}
  \begin{minipage}{0.45\linewidth}
      \centerline{\includegraphics[height=2.8cm]{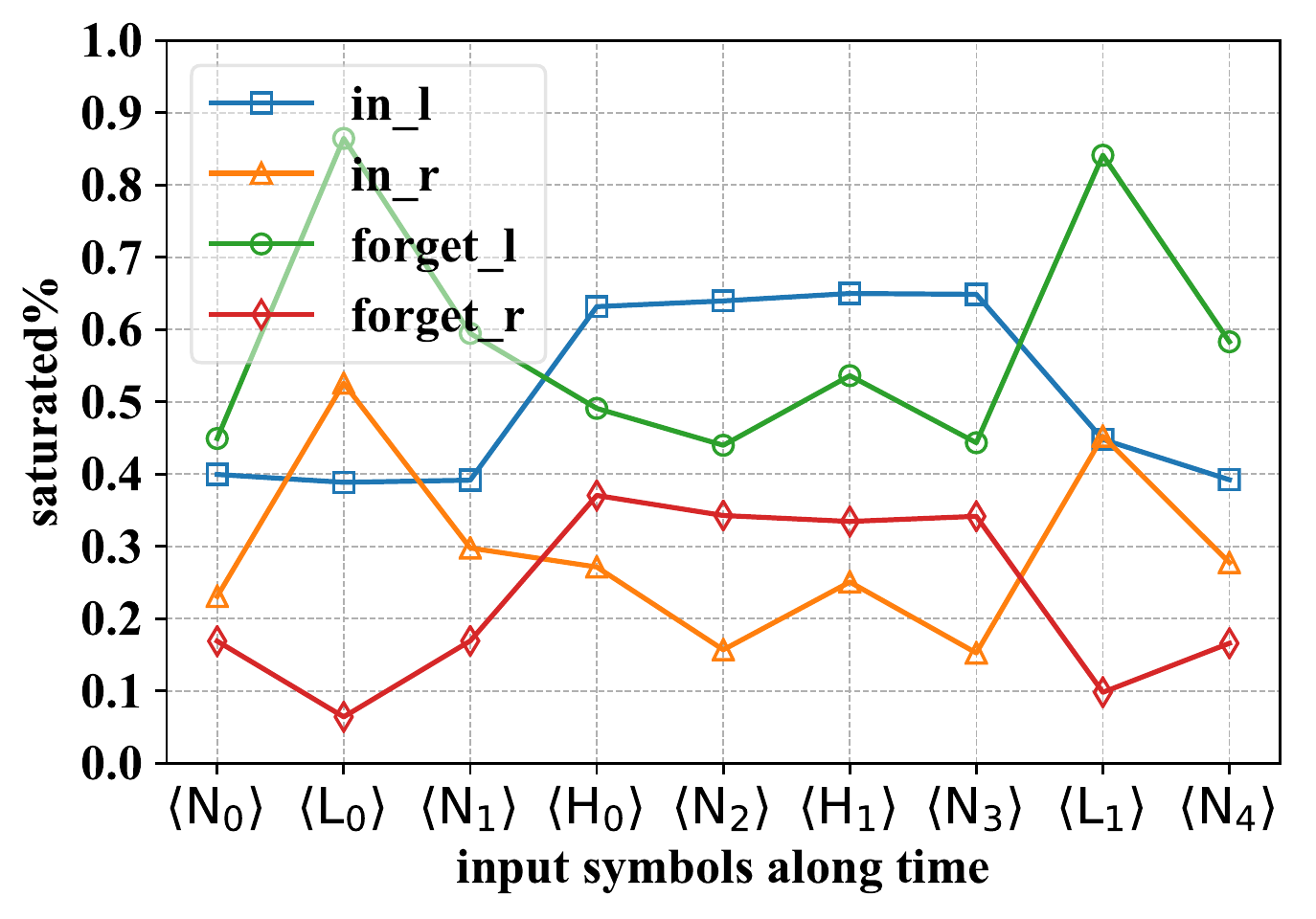}}
      \centerline{a. controller gate}
  \end{minipage}
  &
  \begin{minipage}{0.45\linewidth}
      \centerline{\includegraphics[height=2.8cm]{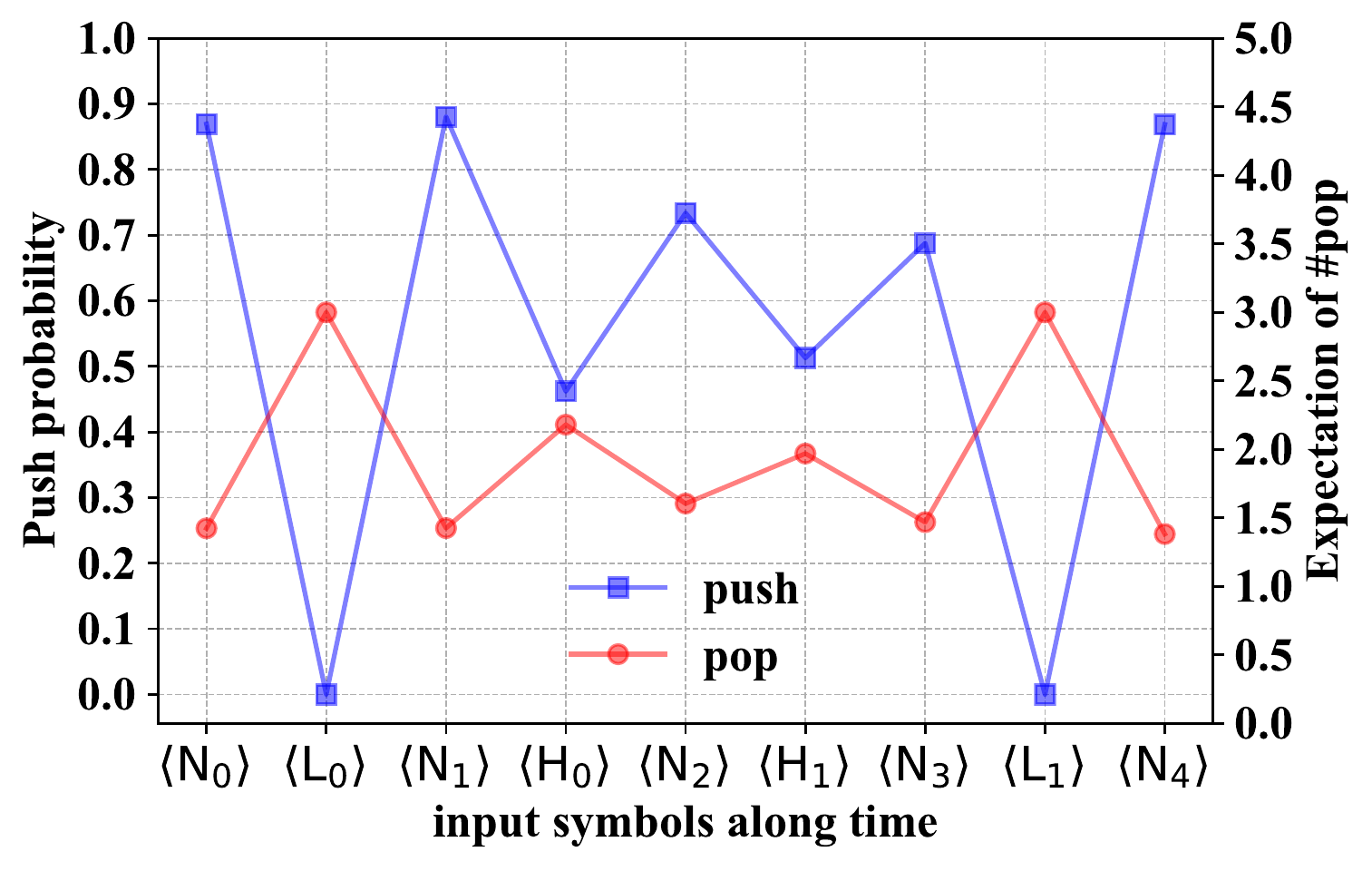}}
      \centerline{b. push-pop policy}
  \end{minipage}
  \end{tabular}
  \caption{Averaged visualization about (a) controller gate and (b) push-pop policy for SANN on \emph{M10AE} task. The x-axis shows each time step represented by categories of input symbol.}
  \label{fig:m10ae_stack_visual}
\end{figure}

In the same way with Section~\ref{sec:analysis_mirror}, we first visualize the controller gates and the read-write  policy of SANN. The results are shown in Figure~\ref{fig:m10ae_stack_visual}. As shown in Figure~\ref{fig:m10ae_stack_visual}a, The controller gates peak at the time steps when the low-priority operators (i.e. $\langle L_0 \rangle$ and $\langle L_1 \rangle$ in Figure~\ref{fig:m10ae_stack_visual}a) appear, which indicates the controller of SANN track specific categories of symbols as the controllers do in mirror task. As shown in Figure~\ref{fig:m10ae_stack_visual}, the push/pop actions are adopted regularly with respect to local structures: 1) every time the lower-priority operator (\emph{+}, \emph{-}) comes, the push probability goes to zero and the expectation of pop times rise up sharply to around $3$; and 2) when dealing with high-priority operations (\emph{*}, \emph{/}), the push/pop lines go up/down with each numeral input relatively more gently. 

\begin{figure}[!htbp]
  \centering
  \includegraphics[width=.45\textwidth]{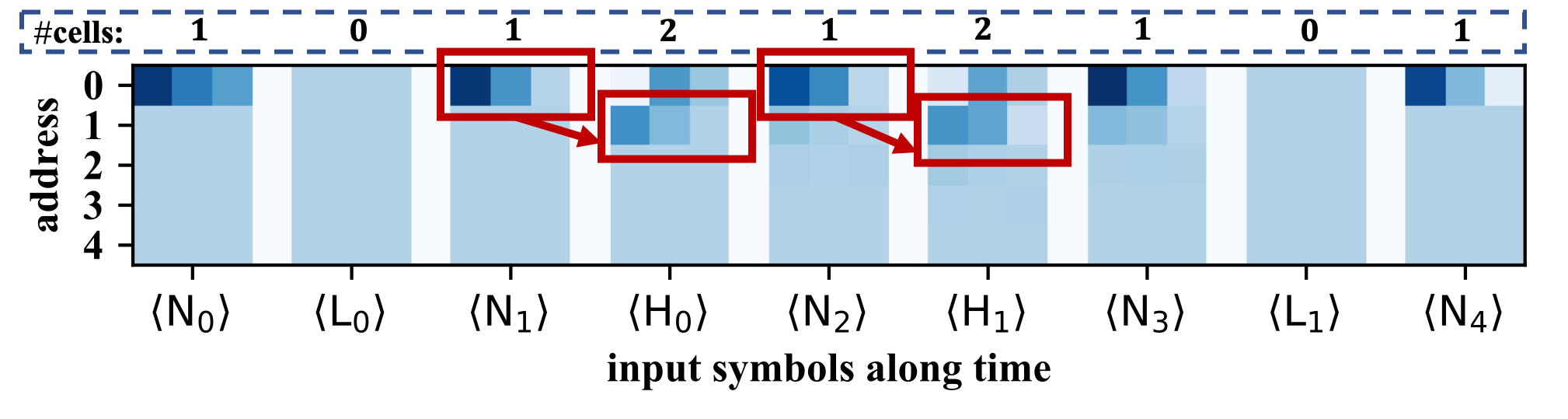}
  \caption{Averaged visualization about memory cells along with input sequence. Each memory cell vector is truncated from $100$-D to its first $3$ dimensions. Darker blue represents higher weight. Blue dashed boxes show the possible number of cells at each time step. Red solid boxes show possible information flow.}
  \label{fig:stack_mem}
\end{figure}

However it is still hard to make a hypothesis about what SANN has learned, and we then visualize the averaged memory at each time step to get further clues shown in Figure~\ref{fig:stack_mem}. The pattern is highly regular: the stack pops all its stored items when it comes to the end of a term (i.e. $\langle {N_0}\rangle$ and $\langle {N_1}\rangle \langle {H_0}\rangle \langle {N_2}\rangle \langle {H_1}\rangle \langle {N_3}\rangle$ here), which is consistent with Figure~\ref{fig:m10ae_stack_visual}b; during processing each term, the stack pushes every time it sees a high-priority operator. 

A hypothesis about what strategy SANN has induced is shown below:

\begin{algorithmic}[1]
\Require 
  arithmetic expression $e=x_0, x_1, \cdots, x_L$
\Ensure
  the evaluation result of $e$

  \State $S \gets [0,0,\cdots,0]^T\in \mathbb R^N$ \Comment{stack}
  \State $l_0$, $l_1$ $\gets$ $0$ \Comment{low/high-priority results}
  \State $l_0'\gets 0$ \Comment{temporary low-priority result}
  \State $p_0$, $p_1$ $\gets$ null \Comment{low/high-priority operators}
  \For{ each input $x_t$ in $e$}
    % \IF{$x_t = {<EOS>}$}
    %   \RETURN $l_0$
    % \ENDIF
    \If{$x_t \in \{+, -\}$} \Comment{for low-priority operators}
      \State $p_0 \gets x_t$ \Comment{save the operator}
      \State pop till empty 
    \ElsIf{$x_t \in \{*, / \}$} \Comment{for high-priority operators}
      \State $l_0 \gets l_0'$ \Comment{adopt the candidate result}
      \State push ($l_1$, $x_t$) \Comment{push the combination}
    \Else \Comment{for numerals}
      \State $l_1, p_1 \gets S[0]$ \Comment{read the combination pushed}
      \State $l_1 \gets \mathrm{eval}(p_1, l_1, x_t)$ \Comment{high-priority operation}
      \State $l_0' \gets l_0$ \Comment{save the candidate result}
      \State $l_0 \gets \mathrm{eval}(p_0, l_0, l_1)$ \Comment{low-priority operation}
      \State pop till empty
      \State push $l_0$ \Comment{push the whole result so far}
    \EndIf
  \EndFor
  \State \Return $l_0$
\end{algorithmic}
where $\mathrm{eval}(f, a_0, a_1)$\footnote{As a special case, if the stack is empty or $f$ is $\mathrm{null}$, the second argument of this function will be returned.} is to evaluate the result of $f(a_0, a_1)$ given the function pointer $f$ and the arguments $a_0$, $a_1$. 

\begin{figure}[!htbp]
  \centering
  \includegraphics[width=.45\textwidth]{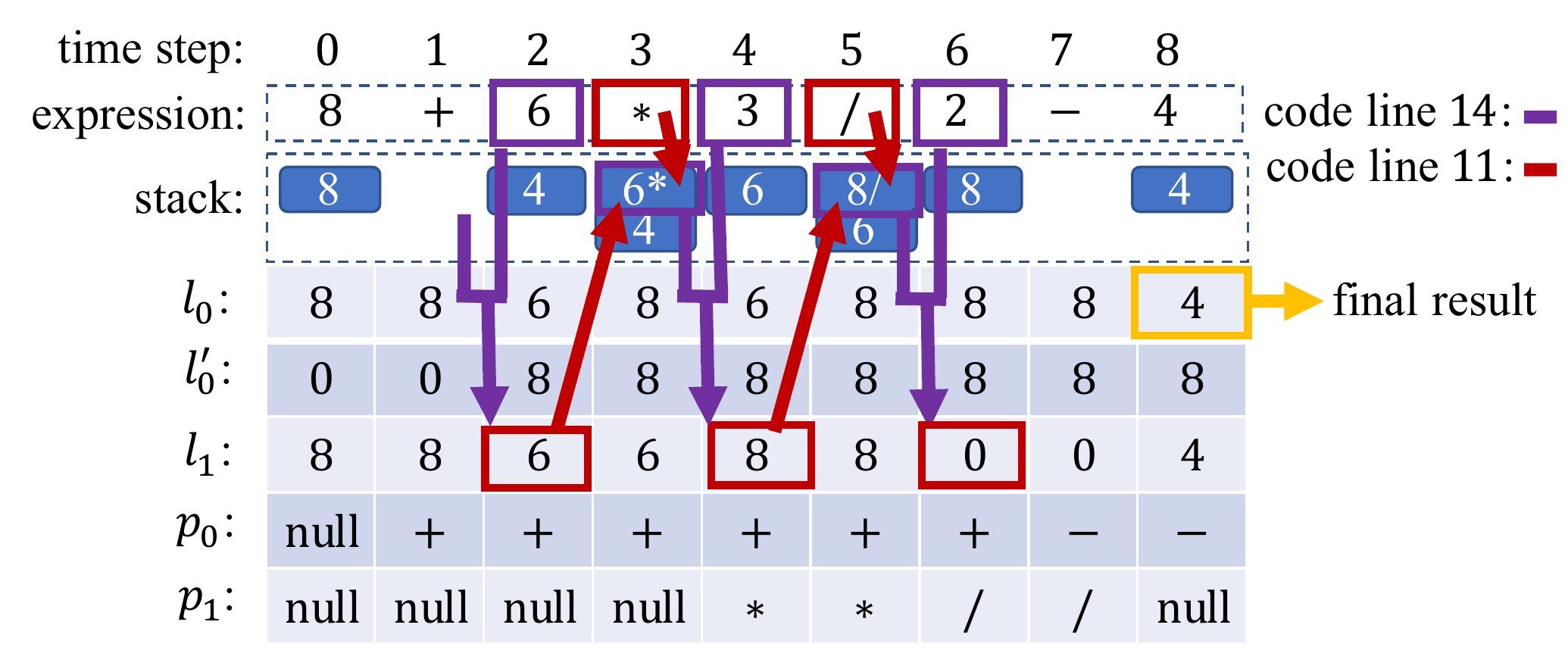}
  \caption{Applying the hypothesized strategy to the example \emph{8+6*3/2-4}. The value of all the variables is displayed at each time step below the expression.}
  \label{fig:stack_procedure}
\end{figure}

There are two kinds of storage in this hypothesis, in-controller storage $l_0, l_0', l_1, p_0, p_1$ and in-memory storage $S$. An example is shown in Figure~\ref{fig:stack_procedure}, where the boxes and arrows in purple (an evaluation step, e.g. \emph{6*3\%10=8} at time step $4$) and red (a combination-pushing step, e.g. push the combination of \emph{6} and \emph{*} at time step $3$) directing the information flow indicates a recursive strategy.

\begin{figure}[!htbp]
  \centering
  \begin{tabular}{ccc}
    \begin{minipage}{0.30\linewidth}
      \centerline{\includegraphics[width=2.5cm]{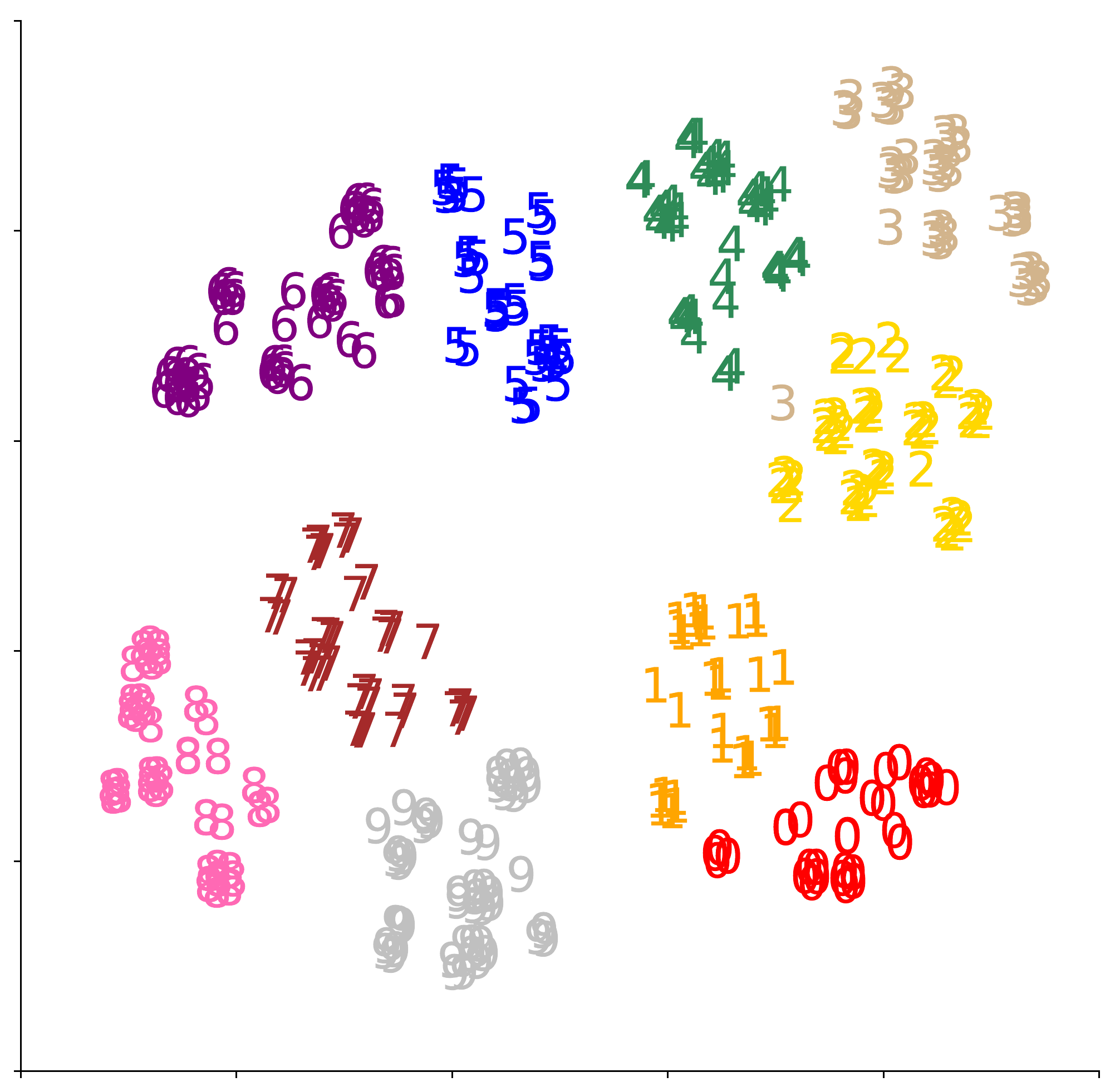}}
      \centerline{\scriptsize a. ($2$, $0$); \tiny $\langle \mathrm{N_0} \rangle \langle \mathrm{L_0} \rangle \langle \mathrm{N_1} \rangle$}

    \end{minipage}
    &
    \begin{minipage}{0.30\linewidth}
      \centerline{\includegraphics[width=2.5cm]{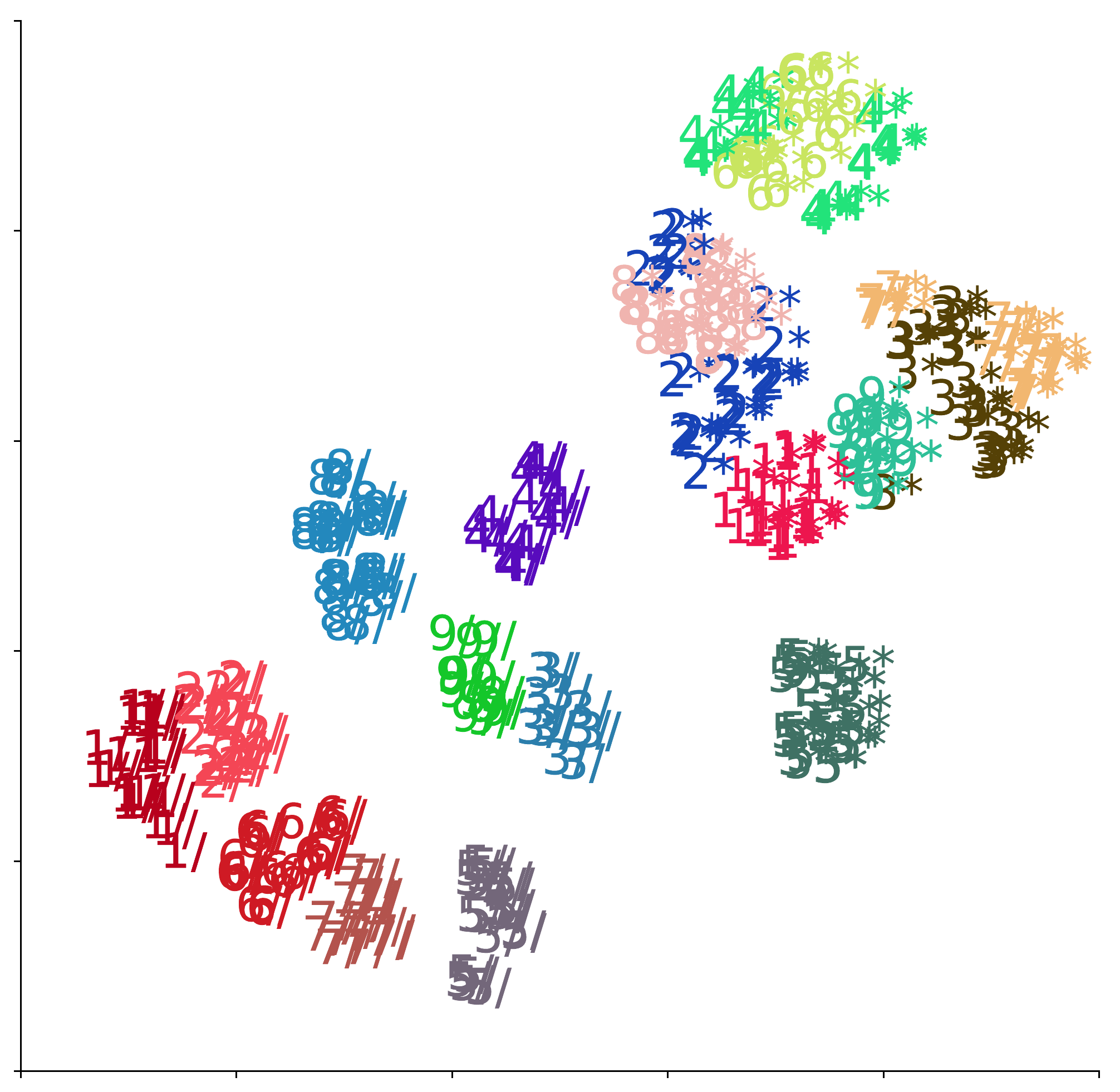}}
      \centerline{\scriptsize b. ($3$, $0$); \tiny $\langle \mathrm{N_1} \rangle \langle \mathrm{H_0} \rangle$}
    \end{minipage}
    &
    \begin{minipage}{0.30\linewidth}
      \centerline{\includegraphics[width=2.5cm]{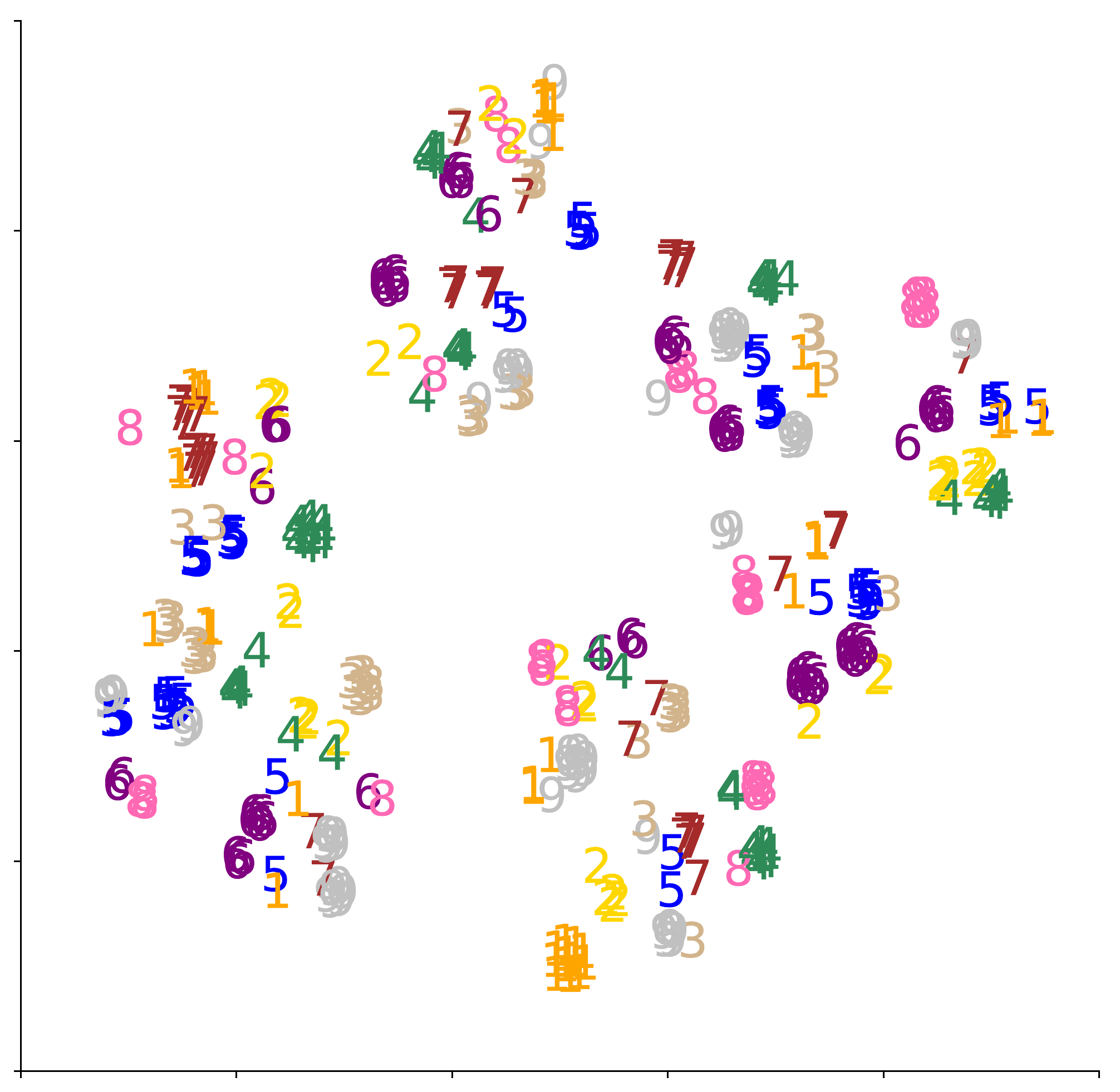}}
      \centerline{\scriptsize c. ($2$, $0$); $\langle \mathrm{N_1} \rangle$}
    \end{minipage}
    
  \end{tabular}
  \caption{Visualization for verifying what the memory cell vectors represent. The parts before the semicolon in the captions are the time step and the memory cell address, noted in the form of ($\left\langle {time\ step} \right\rangle$, $\left\langle {cell\ address} \right\rangle$). And after the semicolon is the hypothesis semantics of the cell vectors, represented as specific parts of $\langle {N_0} \rangle \langle {L_0} \rangle \langle {N_1} \rangle \langle {H_0} \rangle \langle {N_2} \rangle \langle {H_1} \rangle \langle {N_3} \rangle \langle {L_1} \rangle \langle {N_4} \rangle$.}
  \label{fig:verify_m10ae}
\end{figure}

As in Section~\ref{sec:analysis_mirror}, we then verified the hypothesis by our proposed analysis method. The result is shown Figure~\ref{fig:verify_m10ae}, where the clear clusters (in Figure~\ref{fig:verify_m10ae}a and~\ref{fig:verify_m10ae}b) indicate the hypothesis is reasonable. A negative example is also included in Figure~\ref{fig:verify_m10ae}c for illustration about an unreasonable hyothesis.

\section{Related Work}
The related work can be divided into two categories, memory-augmented neural networks and visualization methods.

\subsection{Memory Augmented Neural Network}
The first MANN model is NTM, neural Turing machine~\citep{graves2014neural}, which is proposed to assign logical flow control over external memory to RNNs. These types of models are associated with the automaton theory. The MANNs can be viewed as RNNs with only internal memory augmented with different types of data structures, as a simple DFA is to form complex automaton like PDA, LBA and Turing machine. Thus some work focus on various memory types from different prior bias for specific tasks. For example, an RNN can learn to generalize on context-free grammars augmented with stacks~\citep{joulin2015inferring,grefenstette2015learning}, learn to model shortest syntactic dependence with a gated memory~\citep{gulcehre2017memory} and learn to solve shortest-path tasks with a more general tape memory~\citep{graves2016hybrid}. Some work is dedicated to overcoming the defects of these models, especially for NTMs, such as separating each memory cell into content and address vectors~\citep{gulcehre2016dynamic}, introducing memory allocation and de-allocation protocols~\citep{graves2016hybrid,munkhdalai2017neural}, speeding up the addressing mechanism~\citep{rae2016scaling} and adding adaptive computational design~\citep{yogatama2018memory}.

\subsection{Understanding Recurrent Networks}
Understanding the RNNs and their variants is enjoying renewed interest, as a result of successful applications in a wide range of machine learning problems on sequential data. For one thing, many work focus on what the RNNs remember in their hidden states. Some visualize the dynamics of the LSTM gates in terms of absolute value and saturation rates that keep track of the structure hints~\citep{DBLP:journals/corr/KarpathyJL15,D18-1537}. And some plot t-SNE visualization for clause representations and derivative saliency maps to understand the polarity changes in sentiment analysis~\citep{li2016visualizing}. There are also researches trying training multiple decoders to predict the history inputs for checking what and how much information is stored~\citep{koppula2018understanding}. And in~\cite{W18-5443}, the gradient matrix of the states with respect to the input embeddings is decomposed with SVD to find the principal direction in the input space. For another, which part of the input has a key effect on the model decisions is also a popular topic. Many works view this problem as searching the minimum set of word vectors or their dimensions to flip the models' decision~\citep{DBLP:journals/corr/LiMJ16a,Westhuizen2018Techniques}; and there is other work computing different relevance measures between inputs and outputs to explore the contribution of the words~\citep{W17-5221,ding2017visualizing,van2017visualizing}.

\section{Conclusion}
In this paper, we analyze strategies learned by memory augmented neural networks on task of reversing random sequence and evaluation of arithmetic expressions. By visualizing the controller gates and read-write policy for the memory, we find both models can summarize the input symbols into categories and dynamically change the policy according to these categories. We make hypothesis about what strategy is induced by the models, and verifying them by our proposed novel qualitative analysis method. One can mimic the analysis pipeline for other settings and thus this work helps inspire more researches on the strategy interpretation for MANNs.

\bibliographystyle{named}
\bibliography{ijcai19}

% \section{Saturation rate changes of controller gates}
% \label{app:gates}
% Below are the plots of saturation rate of controller gates in ARNN and TARNN.
% \begin{figure}[H]
%   \centering
%   \begin{tabular}{cc}
%     \begin{minipage}{0.40\linewidth}
%       \centerline{\includegraphics[width=8cm]{f7alstm.png}}
%       \centerline{(a) ARNN}
%     \end{minipage}
%     &
%     \begin{minipage}{0.48\linewidth}
%       \centerline{\includegraphics[width=6.2cm]{f7ntm.png}}
%       \centerline{(b) TARNN}
%     \end{minipage}
%   \end{tabular}
%   \caption{Averaged saturated LSTM gates\% of the controller along with input symbols. `in\_l', `in\_r', `forget\_l' and `forget\_r' represent left saturated input gates, right saturated input gates, left saturated forget gates and right saturated forget gates, respectively.}
%   \label{f7sup}
% \end{figure}

\end{document}